\documentclass{article}

\PassOptionsToPackage{numbers, compress}{natbib}

    \usepackage[preprint]{neurips_2025}

\usepackage[utf8]{inputenc} 
\usepackage[T1]{fontenc}    
\usepackage{hyperref}       
\usepackage{url}            
\usepackage{booktabs}       
\usepackage{amsfonts}       
\usepackage{nicefrac}       
\usepackage{microtype}      
\usepackage{xcolor}         

\usepackage{graphicx}
\usepackage{amsmath}
\usepackage{multirow}
\usepackage{subcaption}
\usepackage{wrapfig}
\usepackage{enumitem}
\usepackage[encapsulated]{CJK}
\usepackage[resetlabels]{multibib}
\newcites{appendix}{References}

\title{CoLa: Chinese Character Decomposition with Compositional Latent Components}

\author{%
  Fan Shi \quad Haiyang Yu \quad Bin Li\thanks{Corresponding author} \quad Xiangyang Xue \\
    Shanghai Key Laboratory of Intelligent Information Processing\\
    School of Computer Science, Fudan University\\
    \texttt{fshi22@m.fudan.edu.cn} \quad \texttt{\{hyyu20,libin,xyxue\}@fudan.edu.cn}
}

\begin{document}

\maketitle

\begin{abstract}
  Humans can decompose Chinese characters into compositional components and recombine them to recognize unseen characters. This reflects two cognitive principles: \textit{Compositionality}, the idea that complex concepts are built on simpler parts; and \textit{Learning-to-learn}, the ability to learn strategies for decomposing and recombining components to form new concepts. These principles provide inductive biases that support efficient generalization. They are critical to Chinese character recognition (CCR) in solving the zero-shot problem, which results from the common long-tail distribution of Chinese character datasets. Existing methods have made substantial progress in modeling compositionality via predefined radical or stroke decomposition. However, they often ignore the learning-to-learn capability, limiting their ability to generalize beyond human-defined schemes. Inspired by these principles, we propose a deep latent variable model that learns \textbf{Co}mpositional \textbf{La}tent components of Chinese characters (CoLa) without relying on human-defined decomposition schemes. Recognition and matching can be performed by comparing compositional latent components in the latent space, enabling zero-shot character recognition. The experiments illustrate that CoLa outperforms previous methods in both character the radical zero-shot CCR. Visualization indicates that the learned components can reflect the structure of characters in an interpretable way. Moreover, despite being trained on historical documents, CoLa can analyze components of oracle bone characters, highlighting its cross-dataset generalization ability.

\end{abstract}

\section{Introduction}
\label{sec:intro}

Humans exhibit remarkable flexibility in recognizing Chinese characters by understanding the internal structure of characters. Chinese characters are composed of components that often carry semantic or categorical cues. Skilled readers can decompose complex characters into constituent parts and generalize across structurally similar but previously unseen characters \cite{shu1997role, chan1998children}. Previous studies show that young children, even English-speaking children, can make informed guesses about unfamiliar Chinese characters based on components \cite{shu2000phonetic, shu2003properties, tang2024compositionality}. Understanding this compositional mechanism offers critical insight into the design of AI systems for Chinese character recognition.

The principles of compositionality and learning-to-learn, widely discussed in cognitive science, have been proposed to explain how humans decompose abstract concepts into constituent parts and recombine known components to form new ones \cite{schyns1998development, winston1975psychology, smith2002object}. Compositionality refers to the idea that complex concepts are structured from simpler parts. Learning-to-learn indicates the ability to automatically acquire strategies for concept decomposition and component recombination. Psychological studies \cite{freyd1983representing} have demonstrated that these principles also underlie the ability to recognize unseen handwritten characters. Inspired by these cognitive mechanisms, some machine learning models \cite{lake2015human, lake2011one, lake2017building} decompose handwritten characters into strokes and recombine them to generate new characters, enabling rapid generalization in simple handwritten characters. These findings suggest that incorporating the principles into intelligence systems is feasible to realize human-like generalization and Chinese character recognition abilities.

\begin{figure*}[!t]
   \centering
   \includegraphics[width=\textwidth]{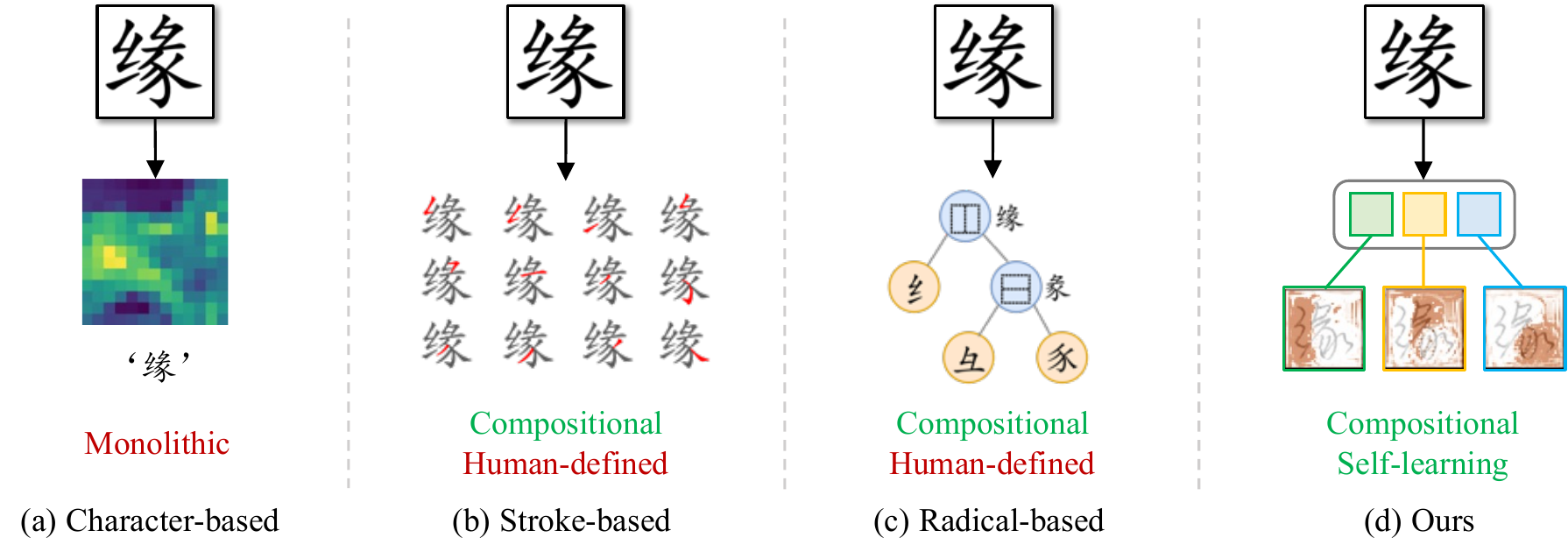}
   \caption{\textbf{Different types of Chinese character recognition methods.} (a) character-based methods extract monolithic representations for prediction; (b) and (c) are stroke-based and radical-based methods requiring human-defined decomposition schemes to predict stroke and radical sequences; (d) the proposed CoLa automatically decomposes characters into compositional components.}
   \label{fig:repr-cmp}
\end{figure*}

Due to the large character set, Chinese character datasets typically follow a long-tail distribution. As a result, the zero-shot recognition problem, where test characters are absent from the training set, is inevitable in practical scenarios, such as the digitalization of Chinese historical documents. The above principles provide inductive biases that support efficient generalization and are critical to Chinese character recognition (CCR) in solving the zero-shot problem. The existing approaches have made substantial progress in modeling compositionality by decomposing characters via predefined schemes. They usually rely on human-defined decomposition rules, for example, as shown in Figure~\ref{fig:repr-cmp}, the stroke- and radical-based approaches~\cite{wang2019radical,wang2018denseran,chen2021zero} utilize an auto-regressive decoder to predict corresponding stroke or radical sequences. Recently, based on CLIP \cite{radford2021learning}, \cite{Yu_2023_ICCV} proposed an efficient image-IDS matching method for zero-shot CCR. However, they often ignore the learning-to-learn capability to automatically acquire decomposition and recombination strategies from data, limiting their ability to generalize beyond predefined decomposition schemes.

Inspired by the compositionality and learning-to-learn principles of human cognition, we propose a deep latent variable model to learn \textbf{Co}mpositional \textbf{La}tent components from Chinese characters (CoLa). CoLa decomposes Chinese characters into latent compositional components without relying on predefined decomposition schemes such as radicals or strokes. CoLa encodes an input image into component-specific representations in a latent space, which are decoded and recombined to reconstruct the visual features of the input. CoLa compares the input image and templates to determine the most likely character class based on the similarity of compositional latent components, which enables zero-shot recognition of Chinese characters. In our experiments, CoLa significantly outperforms previous methods in the radical zero-shot setting. Visualization results further demonstrate that the components learned by CoLa capture the structure of Chinese characters in an interpretable manner. Notably, although trained on historical documents, CoLa can generalize effectively to learn and match components of oracle bone characters, highlighting its cross-dataset generalization ability.

\section{Related Works}
\label{sec:related-work}

\subsection{Zero-shot Chinese Character Recognition}

Due to the significantly larger number of Chinese characters compared to Latin characters, character recognition in Chinese inevitably encounters zero-shot problems, \textit{i.e.}, the characters in the test set are excluded in the training set. Early works in Chinese character recognition can be broadly categorized into three types: \textbf{1) Character-based.} Before the era of deep learning, the character-based methods usually utilize the hand-crafted features to represent Chinese characters~\citep{jin2001study,su2003novel,chang2006techniques}. With deep learning achieving a great success, MCDNN~\citep{cirecsan2015multi} takes the first attempt to use CNN for extracting robust features of Chinese characters while approaching the human performance on handwritten CCR in the ICDAR 2013 competition~\citep{yin2013icdar}. \textbf{2) Radical-based.} To solve the character zero-shot problem, some methods propose to predict the radical sequence of the input character image. In \cite{wang2018denseran}, character images are first fed into a DenseNet-based encoder \cite{huang2017densely} to extract the character features, which are subsequently decoded into the corresponding radical sequences through an attention-based decoder. However, the prediction of radical sequences takes longer time than the character-based methods. Although HDE \cite{cao2020zero} adopts a matching-based method to avoid the time-consuming radical sequence prediction, this method needs to manually design a unique vector for each Chinese character. \textbf{3) Stroke-based.} To fundamentally solve the zero-shot problem, some methods decompose Chinese characters into stroke sequences. The early stroke-based methods usually extract strokes by traditional strategies. For example, in \cite{kim1999decomposition}, the authors employed mathematical morphology to extract each stroke in characters. The proposed method in \cite{liu2001model} describes each Chinese character as an attributed relational graph. Recently, a deep-learning-based method \cite{chen2021zero} is proposed to decompose each Chinese character into a sequence of strokes and employs a feature-matching strategy to solve the one-to-many problem (\textit{i.e.}, there is a one-to-many relationship between stroke sequences and Chinese characters). 

Recently, \cite{Yu_2023_ICCV} introduced CCR-CLIP, which aligns character images with their radical sequences to recognize zero-shot characters, achieving comparable inference efficiency with the character-based approach. All previous methods focus on learning Chinese character features through human-defined representations but struggle to achieve high generalization capabilities.

\subsection{Object-centric Representation Learning}

Object-centric representation methods interpret the world in terms of objects and their relationships. They capture structured representations that are more interpretable, compositional, and generalizable, which has become increasingly popular in computer vision, as it aligns with how humans perceive and interact with the world. One class of models extracts object-centric representations with feedforward processes. For example, SPACE and GNM \cite{lin2020space, jiang2020generative} attempt to divide images into small patches for parallel computation while modeling layouts of scenes. Another class of models initializes and updates object-centric representations by iterative processes \cite{greff2017neural, greff2019multi, emami2021efficient}. A representative method is Slot Attention, which assigns visual features to initialized slots via iterative cross-attention mechanism \cite{locatello2020object}. Based on Slot Attention, many methods have been proposed to improve the quality of object-centric representations in different scenarios \cite{seitzer2022bridging}. Recently, some models have aimed at parsing object-centric scene representations in videos. SAVi++ \cite{elsayed2022savi++} uses Slot Attention to extract a set of temporally consistent latent variables while discovering and segmenting objects with additional visual cues of the first video frame. STEVE \cite{singh2022simple} combines the transformer-based decoder of SLATE \cite{singh2021illiterate} with a standard slot-level recurrence module to extract object-centric representations.

\section{Methodology}
\label{sec:methodology}

In this section, we introduce the proposed CoLa, which is a deep latent variable model to learn compositional latent components from Chinese characters. CoLa encodes Chinese characters into compositional representations in a latent space, capturing the underlying components that constitute these characters. CoLa recognizes zero-shot Chinese characters by matching the input and template images in the latent space. The components are discovered without annotations at the radical or stroke level, nor do they correspond to the predefined standard radicals or strokes, but can illustrate the underlying composition of Chinese characters. In zero-shot CCR, the training and testing character sets contain completely distinct characters. The key to addressing the task lies in generalizing the recognition ability to previously unseen characters. CoLa solves the problem by mapping images into the latent space of compositional components.

\subsection{Overall Framework}

We denote the input character image as $\boldsymbol{X}$ and its class label as $y$, where $y \in \mathcal{C}$ and $\mathcal{C}$ is the set of all class labels. $N$ template character images are provided for each character in $\mathcal{C}$. The set of all template images is $\mathcal{T}$, where $\mathcal{T}_{ij}$ indicates the $j$-th template image of the $i$-th character in $\mathcal{C}$. The templates are typically generated from public font files and matched to determine the class of input character images. Given templates $\mathcal{T}$ and the input image $\boldsymbol{X}$, a traditional CCR model predicts the correct label of $\boldsymbol{X}$ by estimating the likelihood $p(y|\boldsymbol{X},\mathcal{T})$.

\begin{wrapfigure}{r}{0.38\textwidth} 
    \centering
    \includegraphics[width=0.3\textwidth]{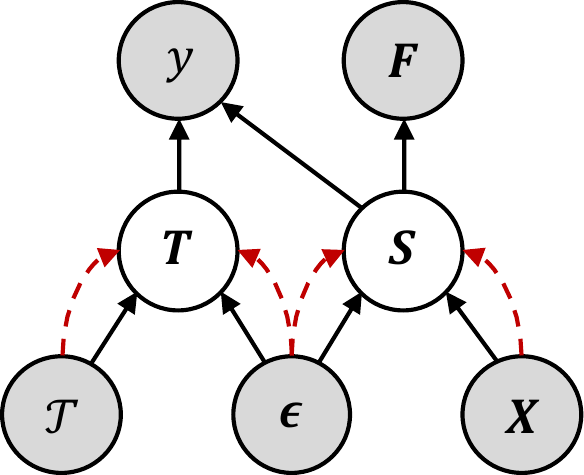}
    \caption{\textbf{The probabilistic graphical model of CoLa.} Gray nodes represent observed variables, while white nodes represent latent variables. Black solid lines indicate the generative process, and red dashed lines indicate the variational distribution.}
   \label{fig:pgm}
\end{wrapfigure}

The probabilistic graphical model of CoLa is shown in Figure \ref{fig:pgm}. CoLa decomposes a Chinese character image into components to extract individual representations in the latent space, which are \textit{compositional latent components}, denoted as $\boldsymbol{S}$ for the input image and $\boldsymbol{T}$ for the template images. CoLa determines the class of $\boldsymbol{X}$ by comparing $\boldsymbol{S}$ and $\boldsymbol{T}$. The compositional latent components reflect the implicit understanding of the Chinese character structures, which are automatically learned by reconstructing the visual features $\boldsymbol{F}$ of the input image, without relying on additional supervised information. $\boldsymbol{S}$ and $\boldsymbol{T}$ are permutation invariant, i.e., using different component orders will not alter the represented character. To model the permutation invariance, CoLa introduces an observed variable $\boldsymbol{\epsilon}$ that specifies the order of components. Instead of defining specific decomposition orders for individual characters, CoLa only requires all images to follow the same order during representation matching. Based on the above definitions, the objective of CoLa is to maximize
\begin{equation}
    \begin{gathered}
        p(\boldsymbol{F},y|\boldsymbol{X},\boldsymbol{\epsilon},\mathcal{T}) = \int p(\boldsymbol{F},y|\boldsymbol{S},\boldsymbol{T},\boldsymbol{X},\boldsymbol{\epsilon},\mathcal{T})p(\boldsymbol{S},\boldsymbol{T}|\boldsymbol{X},\boldsymbol{\epsilon},\mathcal{T}) \mathrm{d}\boldsymbol{S} \mathrm{d}\boldsymbol{T},
    \end{gathered}
    \label{eq:likelihood}
\end{equation}
which is typically intractable when parameterized with complex function approximators like neural networks. The stochastic gradient variational Bayes (SGVB) estimator \cite{kingma2013auto, sohn2015learning} is applied to make Equation \ref{eq:likelihood} tractable by estimating the log-form of the likelihood through the evidence lower bound (ELBO). The core idea is to approximate the posterior distribution $p(\boldsymbol{S},\boldsymbol{T}|\boldsymbol{X},\boldsymbol{\epsilon},\mathcal{T})$ with a variational distribution $q(\boldsymbol{S},\boldsymbol{T}|\boldsymbol{X},\boldsymbol{\epsilon},\mathcal{T},\boldsymbol{F},y)$, whose parameters are modeled by neural networks. Then the log-likelihood is estimated via the following lower bound (see appendix for the detailed derivation):
\begin{equation}
    \begin{gathered}
        \log p(\boldsymbol{F},y|\boldsymbol{X},\boldsymbol{\epsilon},\mathcal{T}) \geq \mathbb{E}_{q(\boldsymbol{S},\boldsymbol{T}|\boldsymbol{X},\mathcal{T},\boldsymbol{\epsilon},\boldsymbol{F},y)} \left[ \log \frac{p(\boldsymbol{F},y,\boldsymbol{S},\boldsymbol{T}|\boldsymbol{X},\boldsymbol{\epsilon},\mathcal{T})}{q(\boldsymbol{S},\boldsymbol{T}|\boldsymbol{X},\mathcal{T},\boldsymbol{\epsilon},\boldsymbol{F},y)} \right] = \text{ELBO}.
    \end{gathered}
    \label{eq:elbo}
\end{equation}
In the following sections, we will introduce the conditional generative process $p(\boldsymbol{F},y,\boldsymbol{S},\boldsymbol{T}|\boldsymbol{X},\boldsymbol{\epsilon},\mathcal{T})$ and the variational distribution $q(\boldsymbol{S},\boldsymbol{T}|\boldsymbol{X},\mathcal{T},\boldsymbol{\epsilon},\boldsymbol{F},y)$ of CoLa, as well as the final form of the ELBO.

\subsection{Conditional Generative Process}

\begin{figure*}[!t]
   \centering
   \includegraphics[width=0.85\textwidth]{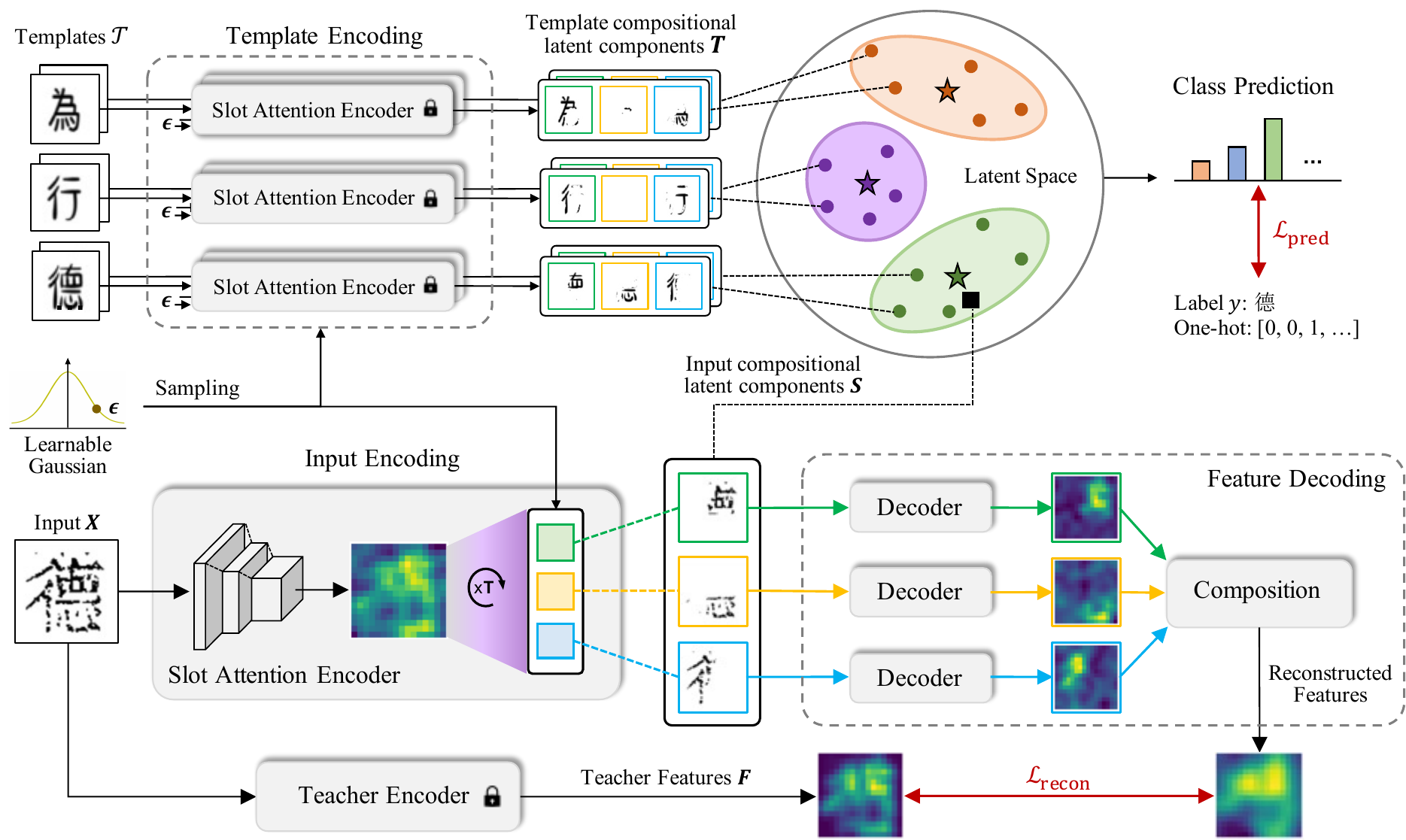}
   \caption{\textbf{The generative process of CoLa.} CoLa extract compositional latent components of the input image and template images in the input encoding anf feature decoding processes, respectively. The input compositional latent components are decoded to reconstruct teacher features. The template compositional latent components constitute a mixture of Gaussians in the latent space, which is used to predict the class of the input image. The training objective is to minimize the distance between the reconstructed features and the features from a frozen teacher encoder, while maximizing the probability that CoLa predicts the correct class label of the input image.}
   \label{fig:overview}
\end{figure*}

According to the generative process illustrated in Figure \ref{fig:pgm}, $p(\boldsymbol{F},y,\boldsymbol{S},\boldsymbol{T}|\boldsymbol{X},\boldsymbol{\epsilon},\mathcal{T})$ is factorized via
\begin{equation}
    p(\boldsymbol{F},y,\boldsymbol{S},\boldsymbol{T}|\boldsymbol{X},\boldsymbol{\epsilon},\mathcal{T}) = \underbrace{p(\boldsymbol{S}|\boldsymbol{X},\boldsymbol{\epsilon})}_{\text{Input Encoding}} \cdot \underbrace{p(\boldsymbol{F}|\boldsymbol{S})}_{\text{Feature Decoding}} \cdot \underbrace{p(\boldsymbol{T}|\mathcal{T},\boldsymbol{\epsilon})}_{\text{Template Encoding}} \cdot \underbrace{p(y|\boldsymbol{S},\boldsymbol{T})}_{\text{Class Prediction}}.
    \label{eq:generative-process}
\end{equation}
CoLa employs an encoder to extract compositional latent components from input images, followed by a decoder to reconstruct visual features based on the representations. The compositional latent components of template images are also extracted and used to predict the class of the input image in the latent space. The generative process of CoLa is composed of the following parts.

\textbf{Input Encoding.} The input image $\boldsymbol{X}$ is downsampled by a CNN-based backbone, augmented with positional embeddings, and transformed into visual features through layer normalization and a multilayer perceptron. We denote the features as $\boldsymbol{H} \in \mathbb{R}^{M \times D}$ where $M$ is the number of input features and $D$ is the dimensionality of each feature. CoLa extracts compositional latent components from the feature using a slot attention module \cite{locatello2020object}, which provides effective inductive biases to cluster similar features in terms of components. We define $p(\boldsymbol{S}|\boldsymbol{X},\boldsymbol{\epsilon})$ as an isotropic Gaussian distribution $\mathcal{N}(\boldsymbol{\mu}^s, \sigma^2 \boldsymbol{I})$, leaving the standard deviation $\sigma$ as a hyperparameter. The mean of the compositional latent components $\boldsymbol{\mu}^s \in \mathbb{R}^{K \times D}$ is initialized by $\boldsymbol{\epsilon}$, where $K$ is the maximum number of components in the images. In the following iterations, the module computes the attention $\boldsymbol{A}$ by measuring the similarity between the components and features, and normalizes the attention to prevent ignoring parts of the input features:
\begin{equation}
   \boldsymbol{A}_{ij} = \frac{e^{\boldsymbol{\Phi}_{ij}}}{\sum_{l} e^{\boldsymbol{\Phi}_{lj}}} ,\quad \text{where } \boldsymbol{\Phi}_{ij} = \frac{f_Q(\boldsymbol{\mu}_i^s) \cdot f_K(\boldsymbol{H}_{j})}{\sqrt{D}}, \quad i=1,\cdots,K,\quad j=1,\cdots,N.
\label{eq:sa-attn}
\end{equation}
Then each feature is assigned to one of $K$ components by a weighted mean operation, and a Gated Recurrent Unit (GRU) fuses the information newly extracted from the features to update $\boldsymbol{\mu}^s$:
\begin{equation}
   \boldsymbol{\mu}_i^s = \text{GRU} \left(\boldsymbol{\mu}_i^s, \sum_{l=1}^{N} \boldsymbol{W}_{il} f_V\left(\boldsymbol{H}_{l}\right) \right), \quad \text{where } \boldsymbol{W}_{ij} = \frac{\boldsymbol{A}_{ij}}{\sum_{l}\boldsymbol{A}_{il}}, \quad i=1,\cdots,K.
\label{eq:sa-upd}
\end{equation}
$f_Q$, $f_K$ and $f_V$ are linear projections. Equations \ref{eq:sa-attn} and \ref{eq:sa-upd} are repeated over multiple iterations, allowing each component to gradually focus on different regions of the image. Controlling $K$ encourages the slot attention module to learn interpretable components rather than decomposing the input into more fragmented parts. Finally, we generate the compositional latent components by $\boldsymbol{S} \sim \mathcal{N}(\boldsymbol{\mu}^s, \sigma^2 \boldsymbol{I})$.

\textbf{Feature Decoding.} The decoder of CoLa transforms the compositional latent components $\boldsymbol{S}$ into the input features $\boldsymbol{F}$ extract by a teacher encoder. $\boldsymbol{H}$ is not chosen as the reconstruction target, since the backbone is updated during training, which may lead the model to exploit shortcuts that minimize reconstruction loss, \textit{e.g.}, collapsing $\boldsymbol{H}$ to a zero matrix. On the other hand, $\boldsymbol{H}$ may contain low-level information to reconstruct image details, while neglecting high-level semantics that are more useful for recognition tasks. Seitzer et al. \cite{seitzer2022bridging} point out that well-pretrained visual features can facilitate the model in learning components that make up images. Inspired by this idea, CoLa introduces a pretrained teacher encoder, aligning the output of the decoder and the teacher encoder to enhance the ability of learning compositional latent components. The teacher encoder consists of a frozen DINOv2 encoder \cite{oquab2023dinov2} and a two-layer convolutional network. The visual features extracted by the teacher are passed through a prediction head, which is jointly trained with the convolutional layers by predicting the classes of character images in the training data. The decoding process $p(\boldsymbol{F}|\boldsymbol{S})$ is formulated as $\mathcal{N}(\boldsymbol{\mu}^d, \sigma^2 \boldsymbol{I})$. CoLa uses a Spatial Broadcast Decoder (SBD) \cite{watters2019spatial} to convert $\boldsymbol{S}_k$ to the component features $\boldsymbol{O}_{k}$ and corresponding mask $\boldsymbol{M}_{k}$: 
\begin{equation}
   \begin{gathered}
      \boldsymbol{M}_{k} = \frac{ e^{\boldsymbol{\Lambda}_{k}}}{\sum_{l=1}^{K} e^{\boldsymbol{\Lambda}_{l}}}, \quad \text{where } \boldsymbol{\Lambda}_{k}, \boldsymbol{O}_{k} = \text{SBD}\left(\boldsymbol{S}_{k}\right), \quad k=1,\cdots,K.
   \end{gathered}
\end{equation}
$\boldsymbol{\Lambda}_{k}$ contains unnormalized logits that indicate where the $k$-th component contributes in the image, and is converted to the normalized mask $\boldsymbol{M}_{k}$. The features of the components are combined via mask-weighted summation, followed by a linear layer to predict the mean $\boldsymbol{\mu}^d = \text{Linear} \left( \sum_{k} \boldsymbol{M}_{k} \odot \boldsymbol{O}_{k} \right)$.

\textbf{Template Encoding.} The template encoding process is factorized in terms of each template image, that is, we have $p(\boldsymbol{T}|\mathcal{T},\boldsymbol{\epsilon}) = \prod_{i \in \mathcal{C}} \prod_{n=1}^{N} p(\boldsymbol{T}_{i,n}|\mathcal{T}_{i,n},\boldsymbol{\epsilon})$. CoLa uses the input encoding process to encode a single template image, where $p(\boldsymbol{T}_{i,n}|\mathcal{T}_{i,n},\boldsymbol{\epsilon}) = \mathcal{N}(\boldsymbol{\mu}^t_{i,n}, \sigma^2 \boldsymbol{I})$ and $\boldsymbol{\mu}^t_{i,n}$ is estimated using the backbone and slot attention module. The compositional latent components $\boldsymbol{T}$ of all template images are acquired by separately sampling $\boldsymbol{T}_{i,n} \sim \mathcal{N}(\boldsymbol{\mu}^t_{i,n}, \sigma^2 \boldsymbol{I})$ for $i \in \mathcal{C}$ and $n = 1,\cdots,N$.

\textbf{Class Prediction.} The input and template images are compared according to $\boldsymbol{S}$ and $\boldsymbol{T}$ in the latent space to predict class labels. $p(\boldsymbol{S}|\boldsymbol{T})$ is regarded as a mixture of Gaussians in the latent space, where the number of mixture components is equal to the size of $\mathcal{C}$ so that each class corresponds to one mixture component. For simplicity, all mixture components have equal weights $1/|\mathcal{C}|$, and each component is an isotropic Gaussian distribution with the covariance $\sigma^2 \boldsymbol{I}$. The mean of the $i$-th mixture component is $\sum_{n=1}^{N} \boldsymbol{T}_{i,n} / N$, which is the center of all $i$-class templates. Predicting the class of an input image is equivalent to identifying which mixture component the input belongs to, which can be described as estimating the posterior of mixture weights:
\begin{equation}
    \begin{gathered}
        p(y=i|\boldsymbol{S},\boldsymbol{T}) =\frac{p(\boldsymbol{S}|\boldsymbol{T},y=i) p(y=i|\boldsymbol{T})}{\sum_{l \in \mathcal{C}} p(\boldsymbol{S}|\boldsymbol{T},y=l) p(y=l|\boldsymbol{T})}
        = \frac{\mathcal{N} \left(\sum_{n=1}^{N} \boldsymbol{T}_{i,n}/N,\sigma^2\boldsymbol{I} \right)}{\sum_{l \in \mathcal{C}} \mathcal{N} \left(\sum_{n=1}^{N} \boldsymbol{T}_{l,n}/N,\sigma^2\boldsymbol{I} \right)}.
    \end{gathered}
    \label{eq:class-prediction}
\end{equation}
Therefore, the class prediction process $p(y|\boldsymbol{S},\boldsymbol{T})$ follows a categorical distribution $\text{Cat}(\boldsymbol{\pi})$, where $\pi_i$ is estimated via Equation \ref{eq:class-prediction}. The class with the highest probability is taken as the final prediction.

\subsection{Variational Inference and Parameter Learning}

In Equation \ref{eq:elbo}, we introduce a variational distribution $q(\boldsymbol{S},\boldsymbol{T}|\boldsymbol{X},\boldsymbol{\epsilon},\mathcal{T},\boldsymbol{F},y)$ as an approximation of the posterior. According to Figure \ref{fig:pgm}, the variational distribution is factorized via
\begin{equation}
   \begin{gathered}
        q(\boldsymbol{S},\boldsymbol{T}|\boldsymbol{X},\boldsymbol{\epsilon},\mathcal{T},\boldsymbol{F},y) = q(\boldsymbol{S}|\boldsymbol{X},\boldsymbol{\epsilon}) 
        q(\boldsymbol{T}|\mathcal{T},\boldsymbol{\epsilon}).
   \end{gathered}
   \label{eq:variational-distribution}
\end{equation}
The variational distribution involves an input encoding and template encoding process, which approximate the posterior of $\boldsymbol{S}$ and $\boldsymbol{T}$, respectively. Both encoding processes take the same forms as in the generative process to extract compositional latent components from the input and templates. Therefore, $q(\boldsymbol{S}|\boldsymbol{X},\boldsymbol{\epsilon})$ and $q(\boldsymbol{T}|\mathcal{T},\boldsymbol{\epsilon})=\prod_{i \in \mathcal{C}} \prod_{n=1}^{N} q(\boldsymbol{T}_{i,n}|\mathcal{T}_{i,n},\boldsymbol{\epsilon})$ are Gaussians with a fixed deviation, and their means are estimated by the slot attention module. We share the backbone and slot attention module in the generative process and variational distribution to reduce the model parameters. Considering Equations \ref{eq:generative-process} and \ref{eq:variational-distribution}, the ELBO in Equation \ref{eq:elbo} is further refined into:
\begin{equation}
    \begin{aligned}
        \text{ELBO} &= \underbrace{\mathbb{E}_{q(\boldsymbol{S}|\boldsymbol{X},\boldsymbol{\epsilon})} \Big[ \log p(\boldsymbol{F}|\boldsymbol{S}) \Big]}_{\text{Reconstruction Term }\mathcal{L}_{\text{recon}}} + \underbrace{\mathbb{E}_{q(\boldsymbol{S},\boldsymbol{T}|\boldsymbol{X},\mathcal{T},\boldsymbol{\epsilon})} \Big[ \log p(y|\boldsymbol{S},\boldsymbol{T}) \Big]}_{\text{Prediction Term }\mathcal{L}_{\text{pred}}} \\
        &\quad\quad\quad\quad - \underbrace{\text{KL}\big(q(\boldsymbol{S}|\boldsymbol{X},\boldsymbol{\epsilon}) \parallel p(\boldsymbol{S}|\boldsymbol{X},\boldsymbol{\epsilon})\big)}_{\text{Input Regularizer }\mathcal{R}_{\text{input}}} - \underbrace{ \text{KL}\big(q(\boldsymbol{T}|\mathcal{T},\boldsymbol{\epsilon}) \parallel p(\boldsymbol{T}|\mathcal{T},\boldsymbol{\epsilon})\big)}_{\text{Template Regularizer }\mathcal{R}_{\text{temp}}}.
    \end{aligned}
    \label{eq:final-elbo}
\end{equation}
The \textbf{reconstruction term} $\mathcal{L}_{\text{recon}}$ encourages the decoder to reconstruct teacher features from compositional latent components, which ensures that the compositional latent components learned by CoLa can reflect the structure of characters. The \textbf{prediction term} $\mathcal{L}_{\text{pred}}$ measures the similarity between the prediction results and ground truth class labels, which directly influences the accuracy of CCR and is critical when training CoLa to recognize Chinese characters. The \textbf{input regularizer} $\mathcal{R}_{\text{input}}$ and the \textbf{template regularizer} $\mathcal{R}_{\text{temp}}$ introduce constraints that the variational distribution should match the posterior distribution. Finally, we compute the ELBO through a Monte Carlo estimator and the training objective of CoLa is to minimize (See appendix for detailed derivations):
\begin{equation}
    \begin{aligned}
        \mathcal{L} &= \Big\| \boldsymbol{F}-\boldsymbol{\tilde{F}} \Big\|^{2}_{2} - \lambda \log \frac{ \exp\left( -\left\|\boldsymbol{\tilde{S}}-\sum_{n=1}^{N} \boldsymbol{\tilde{T}}_{y,n}/N\right\|_{2}^{2} \right)}{\sum_{i \in \mathcal{C}} \exp\left( -\left\|\boldsymbol{\tilde{S}}-\sum_{n=1}^{N} \boldsymbol{\tilde{T}}_{i,n}/N\right\|_{2}^{2} \right)}.
    \end{aligned}
    \label{eq:loss}
\end{equation}
$\boldsymbol{\tilde{F}}$ is the reconstructed features, $\boldsymbol{\tilde{S}}$ and $\boldsymbol{\tilde{T}}$ are latent variables sampled from the variational distribution, and $\lambda$ is a hyperparameter controlling the importance of the prediction term. After training, CoLa can directly recognize zero-shot characters based on the templates of novel character sets.

\section{Experiments}
\label{sec:experiments}

\begin{figure*}[!t]
    \begin{center}
    \includegraphics[width=0.9\textwidth]{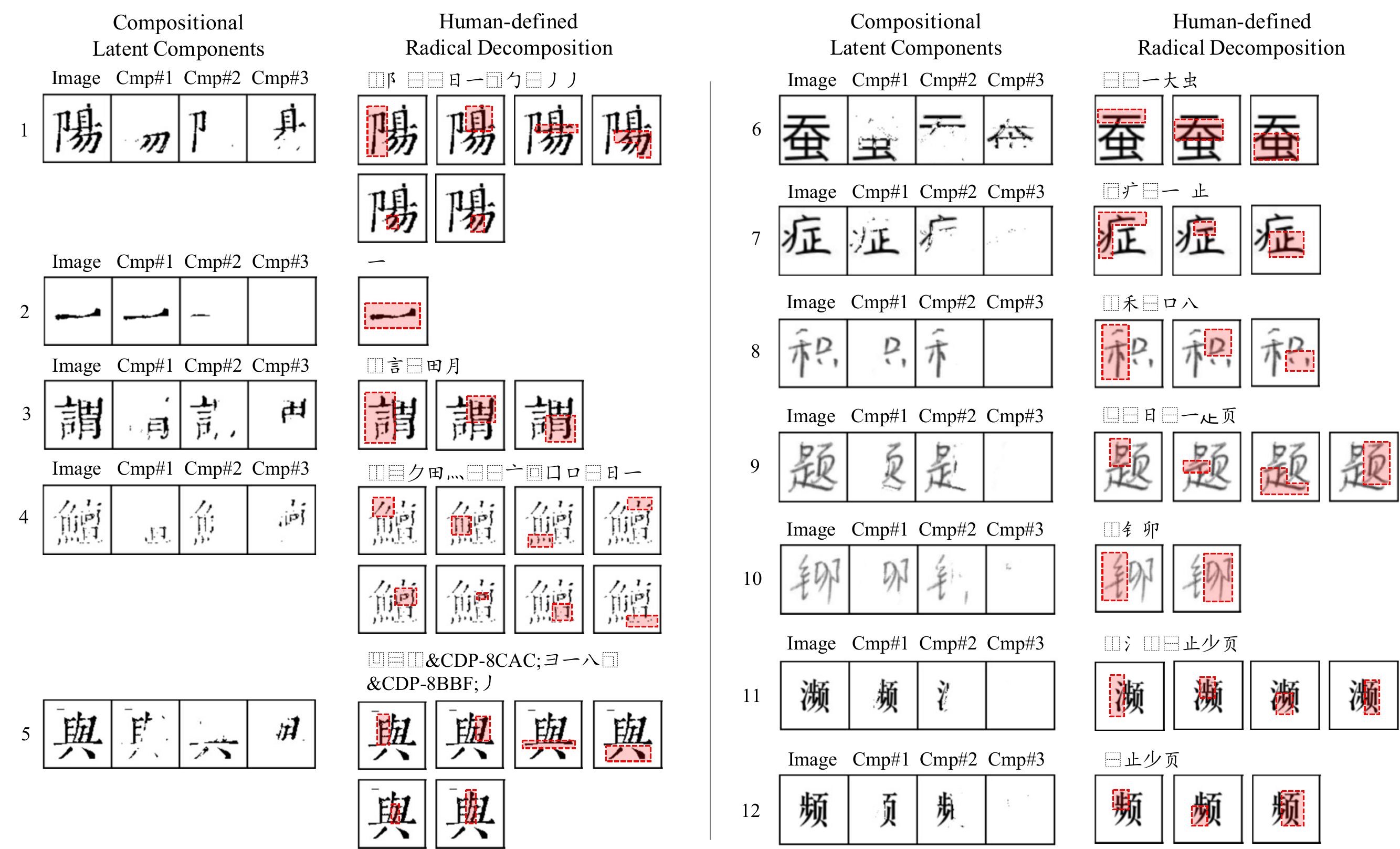}
    \end{center}
    \caption{\textbf{Visualization of the compositional latent components.} For each example, the left side are compositional latent components (Cmp\#1 $\sim$ Cmp\#3), and the right side is the human-defined radical decomposition scheme of the character. The radical regions are highlighted using red boxes.}
    \label{fig:vis_clc}
 \end{figure*}

In this section, we first introduce the experimental settings, including data construction and training details. Then, we show some results of conducted experiments (additional experimental results are shown in the appendix) to illustrate the application of the compositional latent components, including visualization of compositional latent components, zero-shot CCR on three datasets, and an evaluation on oracle bone characters to validate the cross-dataset generalization ability of CoLa.


\textbf{Dataset Construction.} In this paper, we mainly conduct experiments on three datasets: HWDB1.0-1.1 \cite{liu2013online}, Printed artistic characters \cite{chen2021zero} and Historical Documents. HWDB1.0-1.1 \cite{liu2013online} contains 2,678,424 handwritten Chinese character images with 3,881 classes, which is collected from 720 writers and covers 3,755 commonly-used Level-1 Chinese characters. Printed artistic characters \cite{chen2021zero} are generated in 105 font files and contain 394,275 samples for 3,755 Level-1 Chinese characters. Some examples of each dataset are shown in the appendix. We follow \cite{chen2021zero} to construct the corresponding datasets for the character zero-shot and radical zero-shot settings. For the character zero-shot settings, we collect samples with labels falling in the first $m$ classes as the training set and the last $k$ classes as the test set. For the handwritten character dataset HWDB and printed artistic character dataset, $m$ ranges in \{500, 1000, 1500, 2000, 2755\} and $k$ is set to 1000. For the radical zero-shot settings, we first calculate the frequency of each radical in the lexicon. Then the samples of characters that have one or more radicals appearing less than $n$ times are collected as the test set, otherwise, collected as the training set, where $n$ ranges in \{10, 20, 30, 40, 50\} in the radical zero-shot settings.

\textbf{Training Details.} CoLa is trained using the Adam optimizer ~\citep{kingma2014adam} where the momentums $\beta_1$ and $\beta_2$ are set to 0.9 and 0.99. For the CNN-based backbone and slot attention module, we increase the learning rate from 0 to $10^{-4}$ in the first 30K steps and then halve the learning rate every 250K steps. For the spatial broadcast decoder, we increase the learning rate from 0 to $3 \times 10^{-4}$ in the first 30K steps and then halve the learning rate every 250K steps. The training batch size is 32, and the input image of CoLa will be scaled to $80 \times 80$. We set $K=3$, $N=10$ and $\lambda=0.01$. More details of model architecture and training settings are provided in the appendix.

\subsection{Visualization of Compositional Latent Components}
\label{vis_slot_sec}

Previous radical-based or stroke-based methods rely on human-defined decomposition schemes. Radical-based methods suffer from inconsistent decomposition across different characters, which requires the model to extract different features from the same visual characteristics, thereby hindering performance. In addition, stroke-based methods require perceiving fine-grained stroke information, which is challenging for Chinese characters with complex structures. In this section, we visualize the compositional latent components learned by CoLa. As shown in Figure~\ref{fig:vis_clc}, we visualize the regions attended by different components for character samples from Historical Document, Printed and HWDB. The visualization results reveal that each component focuses on distinct and independent regions of the character. Despite the absence of fine-grained supervision information based on radicals or strokes, the compositional latent components can still effectively distinguish meaningful regions of the character. Figure~\ref{fig:vis_clc} shows that CoLa produces component decompositions that, in some cases, align with human-defined schemes (Examples 2, 3, 8, 10). For complex characters from historical documents, humans attempt to split them into a large number of fine-grained elements (Examples 1, 4, 5), whereas CoLa tends to learn higher-level structures. We observe that CoLa performs different hierarchical decompositions for characters. For example, although Comp\#1 of Example 11 can be further decomposed in the same way as Example 12, CoLa chooses to stop at the level of Example 12 without proceeding to finer splits. This suggests that CoLa develops a distinct understanding of decomposition to capture the overall structure of Chinese characters.

\subsection{Results on Chinese Character Recognition}

\begin{table}[!t]
    \centering
    \small
    \caption{\textbf{Accuracy (\%) of Chinese character recognition on the character and radical zero-shot setting.} CoLa outperforms the previous methods on handwritten and printed character datasets HWDB and Printed, especially with a limited training charset (with only 500 training characters). The training of CoLa does not rely on human-defined decomposition schemes, therefore the compositional latent components learned by CoLa can handle zero-shot radicals more effectively.}
    \begin{tabular}{l ccccc ccccc}
    \toprule
    \multirow{2}{*}{Datasets}  & \multicolumn{5}{c}{HWDB (Character Zero-shot)} & \multicolumn{5}{c}{Printed (Character Zero-shot)} \\ 
    \cmidrule(l){2-6} \cmidrule(l){7-11}
    & 500 & 1000 & 1500 & 2000 & 2755 & 500 & 1000 & 1500 & 2000 & 2755 \\
    \midrule
    DenseRAN\cite{wang2018denseran} & 1.70 & 8.44 & 14.71 & 19.51 & 30.68 & 0.20 & 2.26 & 7.89 & 10.86 & 24.80 \\
    HDE\cite{cao2020zero} & 4.90 & 12.77 & 19.25 & 25.13 & 33.49 & 7.48 & 21.13 & 31.75 & 40.43 & 51.41 \\
    SD\cite{chen2021zero} & 5.60 & 13.85 & 22.88 & 25.73 & 37.91 & 7.03 & 26.22 & 48.42 & 54.86 & 65.44 \\
    ACPM\cite{zu2022chinese} & 9.72 & 18.50 & 27.74 & 34.00 & 42.43 & - & - & - & - & -  \\
    CUE\cite{luo2023self} & 7.43 & 15.75 & 24.01 & 27.04 & 40.55 & - & - & - & - & - \\
    SideNet\cite{li2024sidenet} & 5.10 & 16.20 & 33.80 & 44.10 & 50.30 & - & - & - & - & - \\
    HierCode\cite{zhang2025hiercode} & 6.22 & 20.71 & 35.39 & 45.67 & 56.21 & - & - & - & - & - \\
    RSST\cite{yu2024chinese} & 11.56 & 21.83 & 35.32 & 39.22 & 47.44 & 23.12 & 42.21 & 62.29 & 66.86 & 71.32 \\
    
    CCR-CLIP\cite{Yu_2023_ICCV} & 21.79 & 42.99 & 55.86 & 62.99 & 72.98 & 23.67 & 47.57 & 60.72 & 67.34 & 76.44 \\
    Ours & \textbf{68.59} & \textbf{76.58} & \textbf{79.16} & \textbf{81.16} & \textbf{82.71} & \textbf{78.10} & \textbf{85.38} & \textbf{90.32} & \textbf{93.26} & \textbf{92.70} \\
    \midrule
    \multirow{2}{*}{Datasets}  & \multicolumn{5}{c}{HWDB (Radical Zero-shot)} & \multicolumn{5}{c}{Printed (Radical Zero-shot)} \\ 
    \cmidrule(l){2-6} \cmidrule(l){7-11}
    & 50 & 40 & 30 & 20 & 10 & 50 & 40 & 30 & 20 & 10 \\
    \midrule
    DenseRAN\cite{wang2018denseran} & 0.21 & 0.29 & 0.25 & 0.42 & 0.69 & 0.07 & 0.16 & 0.25 & 0.78 & 1.15 \\
    HDE\cite{cao2020zero} & 3.26 & 4.29 & 6.33 & 7.64 & 9.33 & 4.85 & 6.27 & 10.02 & 12.75 & 15.25 \\
    SD\cite{chen2021zero} & 5.28 & 6.87 & 9.02 & 14.67 & 15.83 & 11.66 & 17.23 & 20.62 & 31.10 & 35.81 \\
    ACPM\cite{zu2022chinese} & 4.29 & 6.20 & 7.85 & 10.36 & 12.51 & - & - & - & - & -  \\
    RSST\cite{yu2024chinese} & 7.94 & 11.56 & 15.13 & 15.92 & 20.21 & 13.90 & 19.45 & 26.59 & 34.11 & 38.15 \\
    CCR-CLIP\cite{Yu_2023_ICCV} & 11.15 & 13.85 & 16.01 & 16.76 & 15.96 & 11.89 & 14.64 & 17.70 & 22.03 & 21.27 \\
    Ours & \textbf{70.40} & \textbf{74.80} & \textbf{77.01} & \textbf{80.64} & \textbf{75.78} & \textbf{82.23} & \textbf{84.48} & \textbf{82.20} & \textbf{92.12} & \textbf{94.81} \\
    \bottomrule
    \end{tabular}
    \label{zs_table}
 \end{table}

\begin{table}[t]
    \centering
    \small
    \caption{Accuracy (\%) of Chinese character recognition on historical document characters.}
    \begin{tabular}{l ccccc}
    \toprule
    Models  & Ours & DenseRAN~\cite{wang2018denseran} & SD~\cite{chen2021zero} & CCR-CLIP~\cite{Yu_2023_ICCV} & DMN~\cite{li2020deep} \\ 
    \midrule
    Accuracy & \textbf{57.37} & 13.43 & 11.09 & 22.36 & 31.86 \\
    \bottomrule
    \end{tabular}
    \label{ancient-table}
 \end{table}

We select several radical-based methods~\citep{wang2018denseran,cao2020zero,luo2023self,li2024sidenet,zhang2025hiercode}, stroke-based method~\citep{chen2021zero,yu2024chinese,zu2022chinese} and matching-based method~\citep{Yu_2023_ICCV} as the compared methods in zero-shot settings. For fair comparison, some few-shot CCR models~\citep{li2020deep}, which trained with additional samples from the test character set, are not considered. Moreover, since the character accuracy of character-based methods is almost zero in zero-shot settings, these methods are also not used for comparison.

\textbf{Character Zero-Shot Setting.} We first validate the effectiveness of CoLa in the character zero-shot setting. As shown in Table~\ref{zs_table}, regardless of the handwritten or printed character dataset, the proposed CoLa outperforms previous methods by a clear margin. For instance, in the 500 HWDB character zero-shot setting, the proposed method achieves a performance improvement of about 47\% compared with the previous methods.

\textbf{Radical Zero-Shot Setting.} Following the previous method~\citep{chen2021zero}, we have also conducted corresponding experiments in the radical zero-shot setting. The experimental results shown in Table~\ref{zs_table} indicate that the proposed method achieves the best performance across all sub-settings with an average improvement of about 60\% in accuracy compared to the previous methods. Since we do not introduce manually defined radical or stroke sequences for supervision, CoLa can still achieve satisfying performance in the case of radical zero-shot scenarios. Although the radicals in test sets are not common in training sets, the proposed CoLa can robustly decompose corresponding latent components, improving the zero-shot recognition performance.

\textbf{Historical Document Characters.} We also collected a character dataset from historical documents to validate the effectiveness of our method. The details of constructing the dataset are shown in the appendix. Through the results in Table~\ref{ancient-table}, we observe that the proposed CoLa can achieve a performance improvement of 25.51\% compared with the previous method DMN~\cite{li2020deep}, which validates the robustness of CoLa in more complicated settings.

\subsection{Learning Components of Oracle Bone Characters}

\begin{figure*}[!t]
    \begin{center}
    \includegraphics[width=0.9\textwidth]{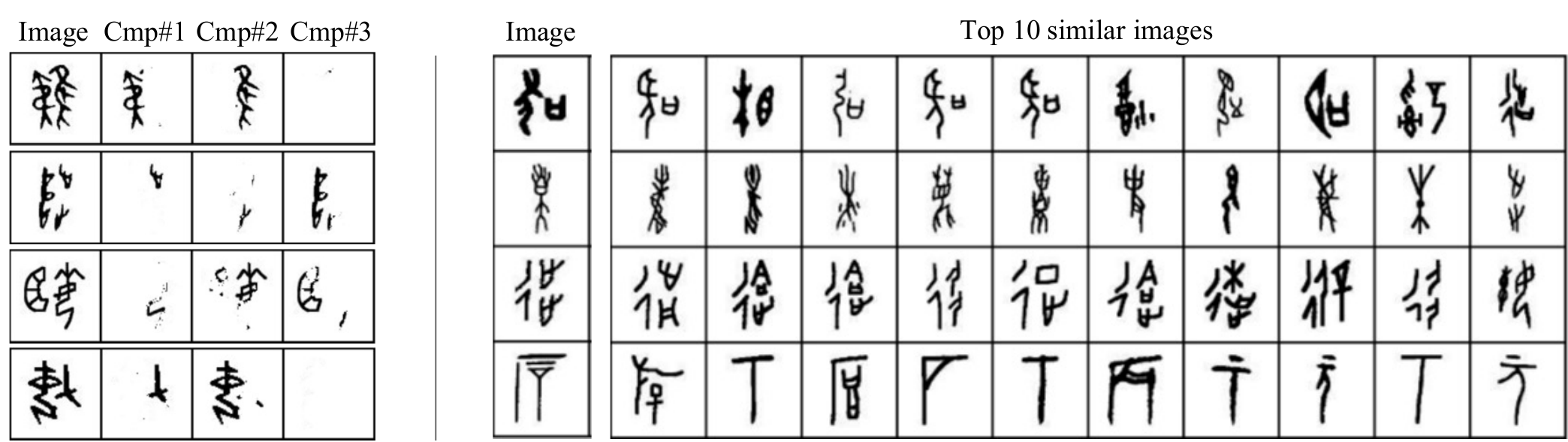}
    \end{center}
    \caption{\textbf{Visualization results on Oracle Bone Characters (OBCs).} Left: we attempt to use CoLa trained on the historical document characters to parse components of OBCs from the HUST-OBC dataset. Right: we select four examples and visualize their top 10 similar samples from 12800 OBCs.}
    \label{vis_obc}
 \end{figure*}

Oracle bone characters (OBCs) are highly stylized and archaic glyphs, which differ from modern Chinese characters in form, making them a valuable benchmark for evaluating the ability to generalize across datasets. To evaluate cross-domain generalization, we apply CoLa, trained on historical document characters, to analyze OBCs from a subset of HUST-OBC \cite{wang2024open}, which contains 12,800 OBCs. As shown on the left panel of Figure \ref{vis_obc}, CoLa can decompose OBCs into interpretable parts, even though it was never trained on OBCs. On the right panel, the top-10 retrievals for each query OBC indicate that CoLa clusters visually and structurally similar characters together. This demonstrates the generalization ability of CoLa to novel types of datasets and highlights its potential for applications in ancient script analysis.

\section{Discussion}
\label{sec:discussion}

In this paper, we propose a deep latent variable model to learn Compositional Latent components of Chinese characters (CoLa) to address challenges in Chinese character recognition, particularly zero-shot recognition. CoLa offers a unique solution by automatically learning compositional components from the data as latent variables, distinct from traditional radical or stroke-based approaches. The experimental results demonstrate that CoLa outperforms previous methods in character and radical zero-shot settings. The visualization experiments also reveal that the acquired components reflect the structure of Chinese characters in an interpretable manner and can be applied to analyze oracle bone characters. We believe that CoLa can offer insights into across-domain recognition applications.

\textbf{Limitation.} Although CoLa achieves outperforming results in zero-shot settings, its capability in scene images with complex backgrounds or low resolution remains underexplored. The complex backgrounds and noise in scene images are unfavorable factors that impact the decomposition of components, which can be a significant topic in future work.



\bibliography{main}
\bibliographystyle{nips}








\newpage
\appendix

\section{Proofs and Derivations}

\subsection{ELBO}

According to the stochastic gradient variational Bayes \citeappendix{kingma2013auto_appendix, sohn2015learning_appendix}, the log-likelihood $p(\boldsymbol{F},y|\boldsymbol{X},\boldsymbol{\epsilon},\mathcal{T})$ can be estimated using the following evidence lower bound (ELBO).
\begin{equation}
    \begin{aligned}
        &\log p(\boldsymbol{F},y|\boldsymbol{X},\boldsymbol{\epsilon},\mathcal{T}) \\
        &= \int q(\boldsymbol{S},\boldsymbol{T}|\boldsymbol{X},\mathcal{T},\boldsymbol{\epsilon},\boldsymbol{F},y) \log p(\boldsymbol{F},y|\boldsymbol{X},\boldsymbol{\epsilon},\mathcal{T}) \mathrm{d}\boldsymbol{S} \mathrm{d}\boldsymbol{T} \\
        &= \int q(\boldsymbol{S},\boldsymbol{T}|\boldsymbol{X},\mathcal{T},\boldsymbol{\epsilon},\boldsymbol{F},y) \log \frac{p(\boldsymbol{F},y,\boldsymbol{S},\boldsymbol{T}|\boldsymbol{X},\boldsymbol{\epsilon},\mathcal{T}) q(\boldsymbol{S},\boldsymbol{T}|\boldsymbol{X},\mathcal{T},\boldsymbol{\epsilon},\boldsymbol{F},y)}{p(\boldsymbol{S},\boldsymbol{T}|\boldsymbol{F},y,\boldsymbol{X},\boldsymbol{\epsilon},\mathcal{T}) q(\boldsymbol{S},\boldsymbol{T}|\boldsymbol{X},\mathcal{T},\boldsymbol{\epsilon},\boldsymbol{F},y)} \mathrm{d}\boldsymbol{S} \mathrm{d}\boldsymbol{T} \\
        &= \int q(\boldsymbol{S},\boldsymbol{T}|\boldsymbol{X},\mathcal{T},\boldsymbol{\epsilon},\boldsymbol{F},y) \log \frac{p(\boldsymbol{F},y,\boldsymbol{S},\boldsymbol{T}|\boldsymbol{X},\boldsymbol{\epsilon},\mathcal{T})}{q(\boldsymbol{S},\boldsymbol{T}|\boldsymbol{X},\mathcal{T},\boldsymbol{\epsilon},\boldsymbol{F},y)} \mathrm{d}\boldsymbol{S} \mathrm{d}\boldsymbol{T} \\
        &\quad\quad\quad\quad + \int q(\boldsymbol{S},\boldsymbol{T}|\boldsymbol{X},\mathcal{T},\boldsymbol{\epsilon},\boldsymbol{F},y) \log \frac{q(\boldsymbol{S},\boldsymbol{T}|\boldsymbol{X},\mathcal{T},\boldsymbol{\epsilon},\boldsymbol{F},y)}{p(\boldsymbol{S},\boldsymbol{T}|\boldsymbol{F},y,\boldsymbol{X},\boldsymbol{\epsilon},\mathcal{T})} \mathrm{d}\boldsymbol{S} \mathrm{d}\boldsymbol{T} \\
        &= \int q(\boldsymbol{S},\boldsymbol{T}|\boldsymbol{X},\mathcal{T},\boldsymbol{\epsilon},\boldsymbol{F},y) \log \frac{p(\boldsymbol{F},y,\boldsymbol{S},\boldsymbol{T}|\boldsymbol{X},\boldsymbol{\epsilon},\mathcal{T})}{q(\boldsymbol{S},\boldsymbol{T}|\boldsymbol{X},\mathcal{T},\boldsymbol{\epsilon},\boldsymbol{F},y)} \mathrm{d}\boldsymbol{S} \mathrm{d}\boldsymbol{T} \\
        &\quad\quad\quad\quad + \text{KL}\big(q(\boldsymbol{S},\boldsymbol{T}|\boldsymbol{X},\mathcal{T},\boldsymbol{\epsilon},\boldsymbol{F},y) \parallel p(\boldsymbol{S},\boldsymbol{T}|\boldsymbol{F},y,\boldsymbol{X},\boldsymbol{\epsilon},\mathcal{T})\big) \\
        &\geq \mathbb{E}_{q(\boldsymbol{S},\boldsymbol{T}|\boldsymbol{X},\mathcal{T},\boldsymbol{\epsilon},\boldsymbol{F},y)} \left[ \log \frac{p(\boldsymbol{F},y,\boldsymbol{S},\boldsymbol{T}|\boldsymbol{X},\boldsymbol{\epsilon},\mathcal{T})}{q(\boldsymbol{S},\boldsymbol{T}|\boldsymbol{X},\mathcal{T},\boldsymbol{\epsilon},\boldsymbol{F},y)} \right] = \text{ELBO}
    \end{aligned}
\end{equation}
Figure \ref{fig:pgm} defines the conditional generative process $p(\boldsymbol{F},y,\boldsymbol{S},\boldsymbol{T}|\boldsymbol{X},\boldsymbol{\epsilon},\mathcal{T})$ and the variational distribution $q(\boldsymbol{S},\boldsymbol{T}|\boldsymbol{X},\mathcal{T},\boldsymbol{\epsilon},\boldsymbol{F},y)$ as in Equations \ref{eq:generative-process} and \ref{eq:variational-distribution}:
\begin{equation}
    \begin{aligned}
        p(\boldsymbol{F},y,\boldsymbol{S},\boldsymbol{T}|\boldsymbol{X},\boldsymbol{\epsilon},\mathcal{T}) &= 
        p(\boldsymbol{S}|\boldsymbol{X},\boldsymbol{\epsilon}) p(\boldsymbol{F}|\boldsymbol{S}) p(\boldsymbol{T}|\mathcal{T},\boldsymbol{\epsilon}) p(y|\boldsymbol{S},\boldsymbol{T}), \\
        q(\boldsymbol{S},\boldsymbol{T}|\boldsymbol{X},\mathcal{T},\boldsymbol{\epsilon},\boldsymbol{F},y) &= q(\boldsymbol{S}|\boldsymbol{X},\boldsymbol{\epsilon}) q(\boldsymbol{T}|\mathcal{T},\boldsymbol{\epsilon}).
    \end{aligned}
\end{equation}
Therefore, the ELBO can be further factorized via Equation \ref{eq:final-elbo}:
\begin{equation}
    \begin{aligned}
        \text{ELBO} &= \mathbb{E}_{q(\boldsymbol{S},\boldsymbol{T}|\boldsymbol{X},\mathcal{T},\boldsymbol{\epsilon},\boldsymbol{F},y)} \left[ \log \frac{p(\boldsymbol{S}|\boldsymbol{X},\boldsymbol{\epsilon}) p(\boldsymbol{F}|\boldsymbol{S}) p(\boldsymbol{T}|\mathcal{T},\boldsymbol{\epsilon}) p(y|\boldsymbol{S},\boldsymbol{T})}{q(\boldsymbol{S}|\boldsymbol{X},\boldsymbol{\epsilon}) q(\boldsymbol{T}|\mathcal{T},\boldsymbol{\epsilon})} \right] \\
        &= \mathbb{E}_{q(\boldsymbol{S}|\boldsymbol{X},\boldsymbol{\epsilon})} \Big[\mathbb{E}_{q(\boldsymbol{T}|\mathcal{T},\boldsymbol{\epsilon})} \Big[ \log p(\boldsymbol{F}|\boldsymbol{S}) \Big] \Big] + \mathbb{E}_{q(\boldsymbol{S}|\boldsymbol{X},\boldsymbol{\epsilon})} \Big[\mathbb{E}_{q(\boldsymbol{T}|\mathcal{T},\boldsymbol{\epsilon})} \Big[ \log p(y|\boldsymbol{S},\boldsymbol{T}) \Big] \Big] \\
        + &\mathbb{E}_{q(\boldsymbol{S}|\boldsymbol{X},\boldsymbol{\epsilon})} \left[\mathbb{E}_{q(\boldsymbol{T}|\mathcal{T},\boldsymbol{\epsilon})} \left[ \log \frac{p(\boldsymbol{S}|\boldsymbol{X},\boldsymbol{\epsilon})}{q(\boldsymbol{S}|\boldsymbol{X},\boldsymbol{\epsilon})} \right] \right] + \mathbb{E}_{q(\boldsymbol{S}|\boldsymbol{X},\boldsymbol{\epsilon})} \left[\mathbb{E}_{q(\boldsymbol{T}|\mathcal{T},\boldsymbol{\epsilon})} \left[ \log \frac{p(\boldsymbol{T}|\mathcal{T},\boldsymbol{\epsilon})}{q(\boldsymbol{T}|\mathcal{T},\boldsymbol{\epsilon})} \right] \right] \\
        & = \underbrace{\mathbb{E}_{q(\boldsymbol{S}|\boldsymbol{X},\boldsymbol{\epsilon})} \Big[ \log p(\boldsymbol{F}|\boldsymbol{S}) \Big]}_{\text{Reconstruction Term }\mathcal{L}_{\text{recon}}} + \underbrace{\mathbb{E}_{q(\boldsymbol{S},\boldsymbol{T}|\boldsymbol{X},\mathcal{T},\boldsymbol{\epsilon})} \Big[ \log p(y|\boldsymbol{S},\boldsymbol{T}) \Big]}_{\text{Prediction Term }\mathcal{L}_{\text{pred}}} \\
        &\quad\quad\quad \quad - \underbrace{\text{KL}\big(q(\boldsymbol{S}|\boldsymbol{X},\boldsymbol{\epsilon}) \parallel p(\boldsymbol{S}|\boldsymbol{X},\boldsymbol{\epsilon})\big)}_{\text{Input Regularizer }\mathcal{R}_{\text{input}}} - \underbrace{\text{KL}\big(q(\boldsymbol{T}|\mathcal{T},\boldsymbol{\epsilon}) \parallel p(\boldsymbol{T}|\mathcal{T},\boldsymbol{\epsilon})\big)}_{\text{Template Regularizer }\mathcal{R}_{\text{temp}}}.
    \end{aligned}
\end{equation}

\subsection{Details of the Conditional Generative Process}

The template encoding process is factorized via $p(\boldsymbol{T}|\mathcal{T},\boldsymbol{\epsilon}) = \prod_{i \in \mathcal{C}} \prod_{n=1}^{N} p(\boldsymbol{T}_{i,n}|\mathcal{T}_{i,n},\boldsymbol{\epsilon})$. Denoting the CNN-based backbone and the slot attention module used in input and template encoding as a function SlotAttnEnc, the compositional latent components of the input image $\boldsymbol{X}$ and templates $\mathcal{T}$ are extracted by
\begin{equation}
    \begin{gathered}
        \boldsymbol{S} \sim \mathcal{N} \left( \boldsymbol{\mu}^{s}, \sigma^2\boldsymbol{I} \right), \quad \text{where } \boldsymbol{\mu}^{s} = \text{SlotAttnEnc} \left( \boldsymbol{X} \right), \\
        \boldsymbol{T}_{i,n} \sim \mathcal{N} \big( \boldsymbol{\mu}^{t}_{i,n},\sigma^2\boldsymbol{I} \big), \quad \text{where } \boldsymbol{\mu}^{t}_{i,n} = \text{SlotAttnEnc} \left( \mathcal{T}_{i,n} \right), \quad i \in \mathcal{C}, \quad n = 1, \cdots, N.
    \end{gathered}
\end{equation}
Denoting the Spatial Broadcast Decoder and the composition process used in feature decoding as a function CompDec, the reconstructed teacher features are obtained via
\begin{equation}
    \begin{gathered}
        \boldsymbol{\tilde{F}} \sim \mathcal{N} \left( \boldsymbol{\mu}^{d}, \sigma^2\boldsymbol{I} \right), \quad \text{where } \boldsymbol{\mu}^{d} = \text{CompDec} \left( \boldsymbol{S} \right).
    \end{gathered}
\end{equation}
In the class prediction process, the final prediction is sampled from a Categorical distribution. The process is $y \sim \text{Cat}(\boldsymbol{\pi})$ where
\begin{equation}
    \begin{gathered}
        \pi_{i} =\frac{\mathcal{N} \left(\sum_{n=1}^{N} \boldsymbol{T}_{i,n} / N,\sigma^2\boldsymbol{I} \right)}{\sum_{l \in \mathcal{C}} \mathcal{N} \left(\sum_{n=1}^{N} \boldsymbol{T}_{l,n} / N,\sigma^2\boldsymbol{I} \right)} = \frac{ \exp\left( -\frac{1}{2\sigma^2} \left\|\boldsymbol{S}-\sum_{n=1}^{N} \boldsymbol{T}_{i,n}/N\right\|_{2}^{2} \right)}{\sum_{l \in \mathcal{C}} \exp\left( -\frac{1}{2\sigma^2} \left\|\boldsymbol{S}-\sum_{n=1}^{N} \boldsymbol{T}_{l,n}/N\right\|_{2}^{2} \right)}.
    \end{gathered}
\end{equation}

\subsection{Details of the Variational Distribution}

The variational distribution shares input and template encoding processes similar to the conditional generative process. The templates are encoded via $q(\boldsymbol{T}|\mathcal{T},\boldsymbol{\epsilon}) = \prod_{i \in \mathcal{C}} \prod_{n=1}^{N} q(\boldsymbol{T}_{i,n}|\mathcal{T}_{i,n},\boldsymbol{\epsilon})$, and the compositional latent components are extracted by
\begin{equation}
    \begin{gathered}
        \boldsymbol{\tilde{S}} \sim \mathcal{N} \left( \boldsymbol{\tilde{\mu}}^{s}, \sigma^2\boldsymbol{I} \right), \quad \text{where } \boldsymbol{\tilde{\mu}}^{s} = \text{SlotAttnEnc} \left( \boldsymbol{X} \right), \\
        \boldsymbol{\tilde{T}}_{i,n} \sim \mathcal{N} \big( \boldsymbol{\tilde{\mu}}^{t}_{i,n},\sigma^2\boldsymbol{I} \big), \quad \text{where } \boldsymbol{\tilde{\mu}}^{t}_{i,n} = \text{SlotAttnEnc} \left( \mathcal{T}_{i,n} \right), \quad i \in \mathcal{C}, \quad n = 1, \cdots, N.
    \end{gathered}
    \label{eq:appendix-variational-distribution}
\end{equation}

\subsection{Monte Carlo Estimator of the ELBO}

Using the detailed definition of the conditional generative process and variational distribution, the terms in the ELBO can be estimated by a Monte Carlo estimator as follows.
\begin{equation}
    \begin{gathered}
        \mathcal{L}_{\text{recon}} = \mathbb{E}_{q(\boldsymbol{S}|\boldsymbol{X},\boldsymbol{\epsilon})} \Big[ \log p(\boldsymbol{F}|\boldsymbol{S}) \Big] \approx - \frac{1}{2\sigma^2} \Big\| \boldsymbol{F} - \text{CompDec}(\boldsymbol{\tilde{S}}) \Big\|^{2}_{2} + C(\sigma), \\
        \mathcal{L}_{\text{pred}} = \mathbb{E}_{q(\boldsymbol{S},\boldsymbol{T}|\boldsymbol{X},\mathcal{T},\boldsymbol{\epsilon})} \Big[ \log p(y|\boldsymbol{S},\boldsymbol{T}) \Big] \approx \log \frac{ \exp\left( -\frac{1}{2\sigma^2} \left\|\boldsymbol{\tilde{S}}-\sum_{n=1}^{N} \boldsymbol{\tilde{T}}_{y,n}/N\right\|_{2}^{2} \right)}{\sum_{l \in \mathcal{C}} \exp\left( -\frac{1}{2\sigma^2} \left\|\boldsymbol{\tilde{S}}-\sum_{n=1}^{N} \boldsymbol{\tilde{T}}_{l,n}/N\right\|_{2}^{2} \right)},
    \end{gathered}
\end{equation}
where $C(\sigma)$ is a constant related to the standard deviation $\sigma$, $\boldsymbol{\tilde{S}}$ and $\boldsymbol{\tilde{T}}$ are compositional latent components sampled from the variational distribution through Equation \ref{eq:appendix-variational-distribution}. Since the conditional generative process and the variational distribution share the same backbone and slot attention module in image encoding and template encoding, we have $p(\boldsymbol{S}|\boldsymbol{X},\boldsymbol{\epsilon})=q(\boldsymbol{S}|\boldsymbol{X},\boldsymbol{\epsilon})$ and $p(\boldsymbol{T}_{i,n}|\mathcal{T}_{i,n},\boldsymbol{\epsilon})=q(\boldsymbol{T}_{i,n}|\mathcal{T}_{i,n},\boldsymbol{\epsilon})$ if the same input and templates are given. Then, the two regularizers are 
\begin{equation}
    \begin{gathered}
        \mathcal{R}_{\text{input}} = \text{KL}\big(q(\boldsymbol{S}|\boldsymbol{X},\boldsymbol{\epsilon}) \parallel p(\boldsymbol{S}|\boldsymbol{X},\boldsymbol{\epsilon})\big) = 0, \\
        \mathcal{R}_{\text{temp}} = \text{KL}\big(q(\boldsymbol{T}|\mathcal{T},\boldsymbol{\epsilon}) \parallel p(\boldsymbol{T}|\mathcal{T},\boldsymbol{\epsilon})\big) = \sum_{i \in \mathcal{C}} \sum_{n=1}^{N} \text{KL}\big(q(\boldsymbol{T}_{i,n}|\mathcal{T}_{i,n},\boldsymbol{\epsilon}) \parallel p(\boldsymbol{T}_{i,n}|\mathcal{T}_{i,n},\boldsymbol{\epsilon})\big) = 0.
    \end{gathered}
\end{equation}
The Monte Carlo estimation of the ELBO is given by
\begin{equation}
    \begin{aligned}
        \text{ELBO}_{MC} &= - \frac{1}{2\sigma^2} \Big\| \boldsymbol{F} - \text{CompDec}(\boldsymbol{\tilde{S}}) \Big\|^{2}_{2} + \log \frac{ \exp\left( -\frac{1}{2\sigma^2} \left\|\boldsymbol{\tilde{S}} - \frac{\sum_{n=1}^{N} \boldsymbol{\tilde{T}}_{c,n}}{N}\right\|_{2}^{2} \right)}{\sum_{l \in \mathcal{C}} \exp\left( -\frac{1}{2\sigma^2} \left\|\boldsymbol{\tilde{S}} - \frac{\sum_{n=1}^{N} \boldsymbol{\tilde{T}}_{l,n}}{N}\right\|_{2}^{2} \right)}.
    \end{aligned}
\end{equation}
In the loss function (Equation \ref{eq:loss}) of CoLa, we set $\sigma=\sqrt{2}/2$ to simplify the $\text{ELBO}_{MC}$ while introducing a hyperparameter $\lambda$ to balance the importance of different terms:
\begin{equation}
    \begin{aligned}
        \mathcal{L} &= \Big\| \boldsymbol{F} - \text{CompDec}(\boldsymbol{\tilde{S}}) \Big\|^{2}_{2} - \lambda \log \frac{ \exp\left( -\left\|\boldsymbol{\tilde{S}} - \sum_{n=1}^{N} \boldsymbol{\tilde{T}}_{c,n}/N\right\|_{2}^{2} \right)}{\sum_{l \in \mathcal{C}} \exp\left( - \left\|\boldsymbol{\tilde{S}} - \sum_{n=1}^{N} \boldsymbol{\tilde{T}}_{l,n}/N\right\|_{2}^{2} \right)}.
    \end{aligned}
\end{equation}

\section{Datasets}

\begin{figure*}[h]
   \begin{center}
   \includegraphics[width=\textwidth]{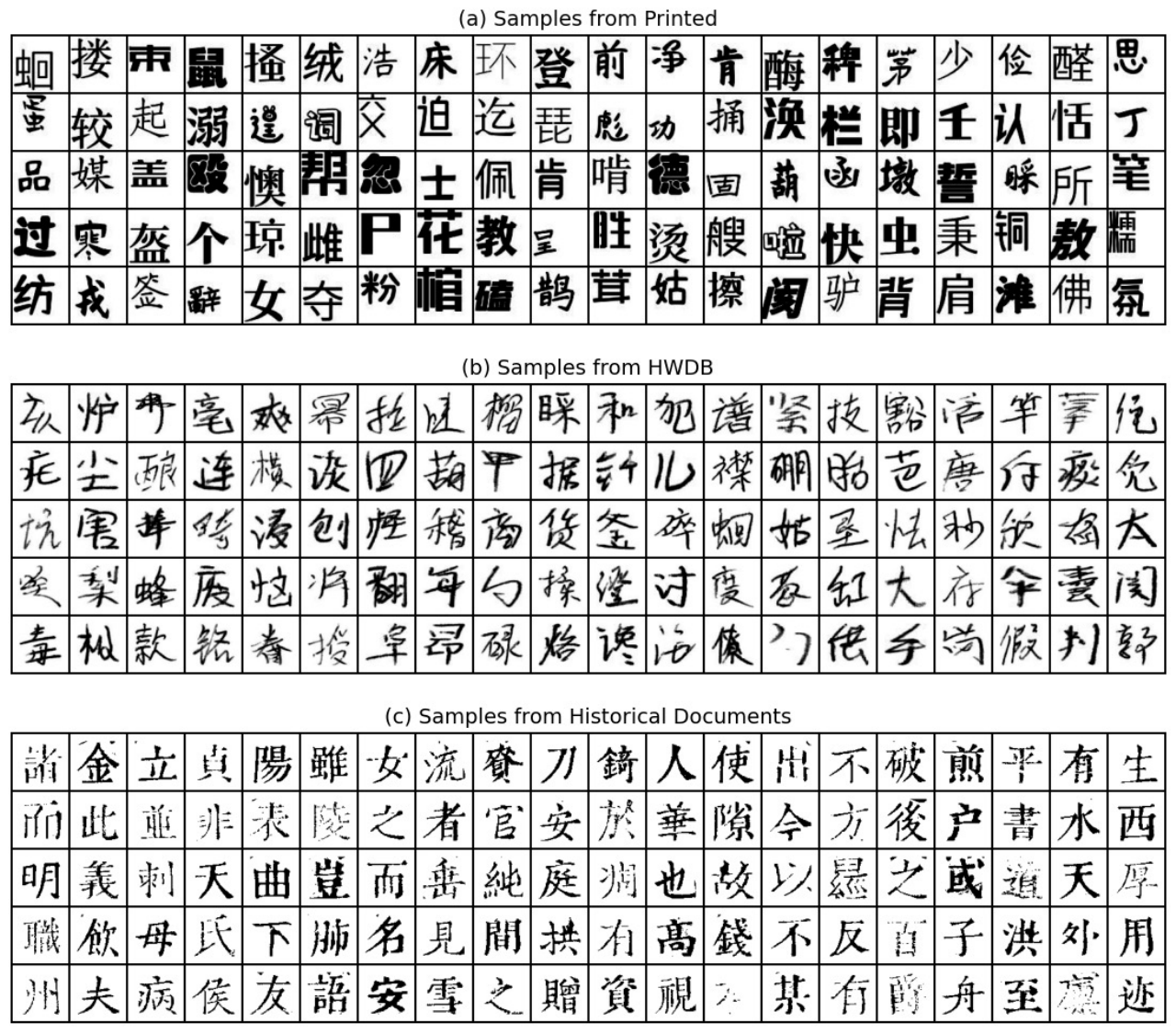}
   \end{center}
   \caption{Visualization of the samples from (a) Printed, (b) HWDB and (c) Historical Documents.}
   \label{fig:appendix-datasets}
\end{figure*}

Figure \ref{fig:appendix-datasets} visualizes the data used in the experiments. The Printed dataset is constructed on the basis of different printed font files. The handwritten dataset HWDB includes handwritten samples from various writers. The template set $\mathcal{T}$ is constructed by selecting $N=10$ images for each character in the character set during the training and testing phases. The templates are generated from commonly used printed fonts to ensure that some rare characters also have templates. The character images in historical documents are collected from the web library\footnote{https://library.harvard.edu/policy-access-digital-reproductions-works-public-domain}. In addition, all character images are allowed to used for free without any copyright concerns.

\section{Details of the Models}

The compared models follow the original experimental settings and architectures. We use open-source codes if the reimplemented methods have official codes (\textit{e.g.}, CCR-CLIP and SD). CoLa is trained on the server with Intel(R) Xeon(R) Gold 6326 CPUs, 24GB NVIDIA GeForce RTX 4090 GPUs, 256GB RAM, and Ubuntu 20.04.6 LTS. CoLa is implemented with PyTorch \citeappendix{paszke2019pytorch_appendix}. 

In the following, we will describe the architectures of learnable networks in CoLa and the hyperparameter selection of CoLa. The learnable networks include: (1) the CNN-based backbone; (2) $f_Q$, $f_K$, $f_V$ and the Gated Recurrent Unit (GRU) in the slot attention module; (3) the Spatial Broadcast Decoder (SBD) and the following linear layer to compose the components; (4) the two-layer CNN and prediction head of the teacher encoder. The details of the learnable networks are provided below.
\begin{itemize}[leftmargin=0.5cm]
  \item \textbf{The CNN-based backbone}:
  \begin{itemize} 
      \item 5 $\times$ 5 Conv, stride 2, padding 2, 192, ReLU
      \item 5 $\times$ 5 Conv, stride 1, padding 2, 192, ReLU
      \item 5 $\times$ 5 Conv, stride 1, padding 2, 192, ReLU
      \item 5 $\times$ 5 Conv, stride 1, padding 2, 192
      \item Cartesian Positional Embedding, 192, LayerNorm
      \item Fully Connected, 192, ReLU
      \item Fully Connected, 192
  \end{itemize}
\end{itemize}
The architecture of the slot attention module follows the original design, which can refer to \citeappendix{locatello2020object_appendix}. Here, we set the slot size (i.e., component size) as 128 and the iteration step as 3. The components have the shape of $K \times 128$, initialized with the $\boldsymbol{\epsilon}$ sampled from a learnable Gaussian $\mathcal{N}(\boldsymbol{\mu}_{\epsilon}, \boldsymbol{\sigma}^2_{\epsilon})$.
\begin{itemize}[leftmargin=0.5cm]
  \item \textbf{SBD}:
  \begin{itemize}
      \item Fully Connected, 192
      \item Learnable 2D Positional Embedding, 192
      \item Fully Connected, 1024, ReLU
      \item Fully Connected, 1024, ReLU
      \item Fully Connected, 1024, ReLU
      \item Fully Connected, 1025
  \end{itemize}
  \item \textbf{The composition linear layer after SBD}:
  \begin{itemize}
      \item Fully Connected, 1024, without bias
  \end{itemize}
\end{itemize}
The SBD output has the shape of $1025 \times 16 \times 16$, split along the channel dimension into the mask of $1 \times 16 \times 16$ and the component feature of $1024 \times 16 \times 16$. Here, $1024 \times 16 \times 16$ is the feature size of the teacher encoder.
\begin{itemize}[leftmargin=0.5cm]
  \item \textbf{The two-layer CNN of the teacher encoder}:
  \begin{itemize}
      \item 3 $\times$ 3 Conv, stride 1, padding 1, 1024, ReLU
      \item 3 $\times$ 3 Conv, stride 1, padding 1, 1024
  \end{itemize}
\end{itemize}
The two-layer CNN converts the DINOv2 features of $768 \times 16 \times 16$ to the teacher features of $1024 \times 16 \times 16$. The teacher encoder is followed by a Transformer encoder block with a class token to predict the label of the input image. The DINOv2 encoder is frozen, and the two-layer CNN and prediction head are trained via a cross-entropy loss based on the training characters and the class labels. Finally, we use the output of the two-layer CNN as the teacher encoder, and the prediction head is not used in the following stages.

We train the teacher encoder using an Adam optimizer \citeappendix{kingma2014adam_appendix}, setting the learning rate to $3\times10^{-4}$ and the batch size to 8. Then we freeze the teacher encoder to train the remaining part of CoLa. We set the learning rate to $3\times10^{-4}$ and the batch size to 32. We first warm up CoLa by disabling the prediction term (i.e., setting $\lambda=0$). After the training loss is stable, we enable the prediction term by setting the $\lambda=0.01$ to train the model. The template images in the training character set are also used to train CoLa in the training process. The CNN-based backbone and the slot attention module are frozen when extracting components of templates to stop the gradient propagation to reduce the cost of computational resources.

\section{Additional Experimental Results}

\subsection{Additional Results of Component Visualization}

\begin{figure*}[!h]
  \centering
  \begin{subfigure}[b]{\textwidth}
      \centering
      \includegraphics[width=0.95\textwidth]{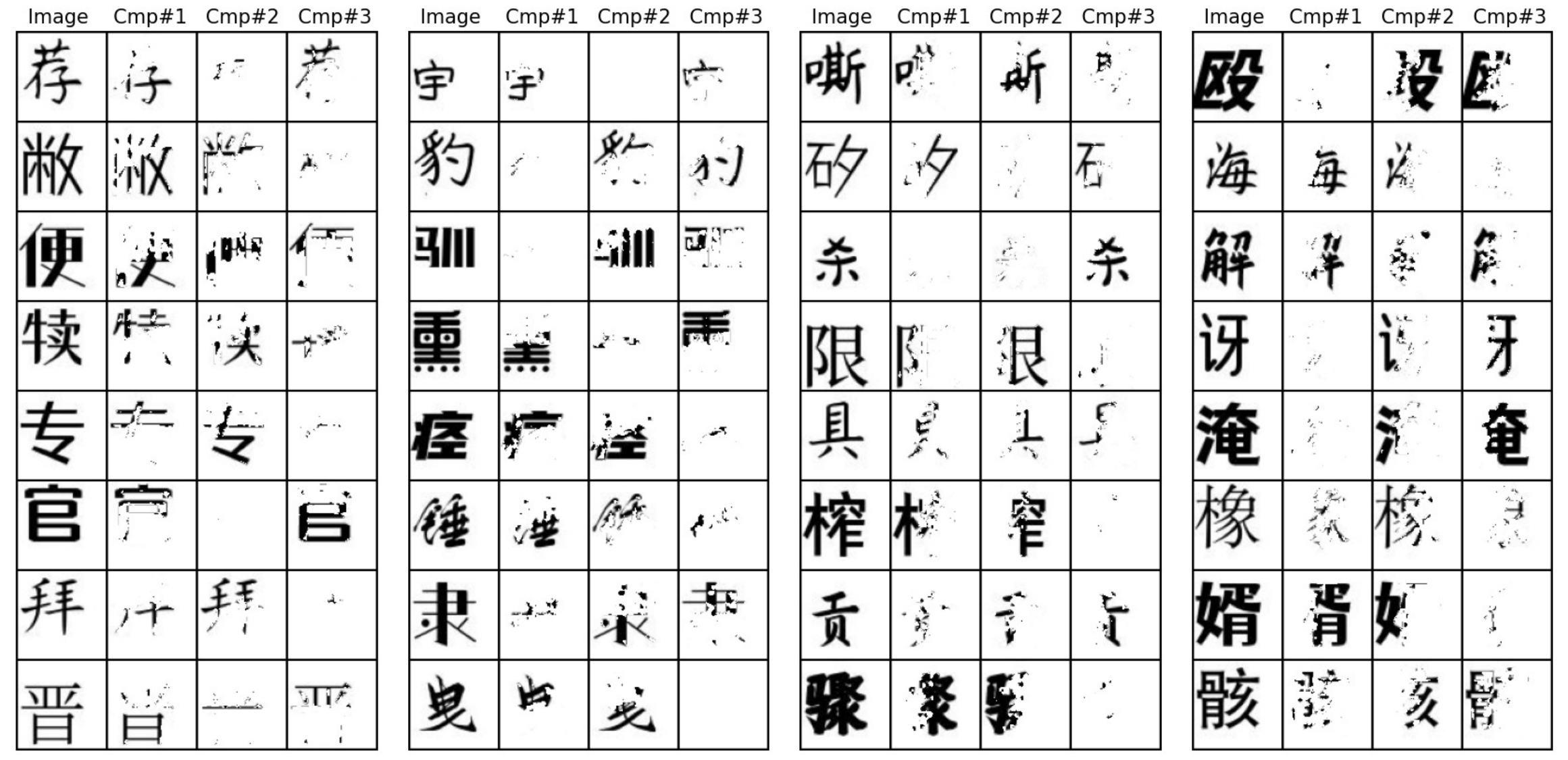}
      \caption{Printed}
      \label{fig:appendix-comp-vis-print}
  \end{subfigure}
  \vskip 0.2cm
  \vfill
  \begin{subfigure}[b]{\textwidth}
      \centering
      \includegraphics[width=0.95\textwidth]{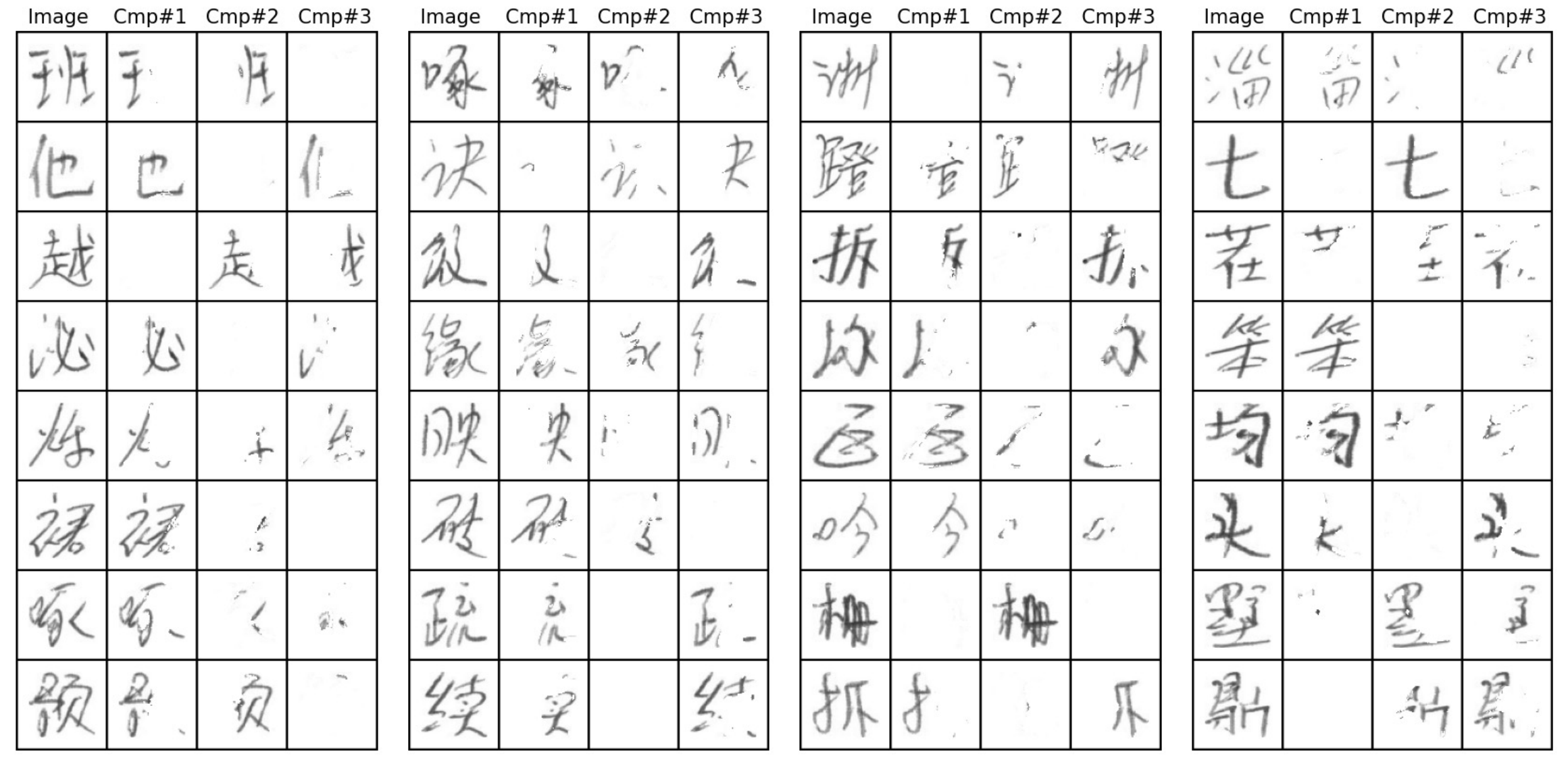}
      \caption{HWDB}
      \label{fig:appendix-comp-vis-hwdb}
  \end{subfigure}
  \vskip 0.2cm
  \vfill
  \begin{subfigure}[b]{\textwidth}
      \centering
      \includegraphics[width=0.95\textwidth]{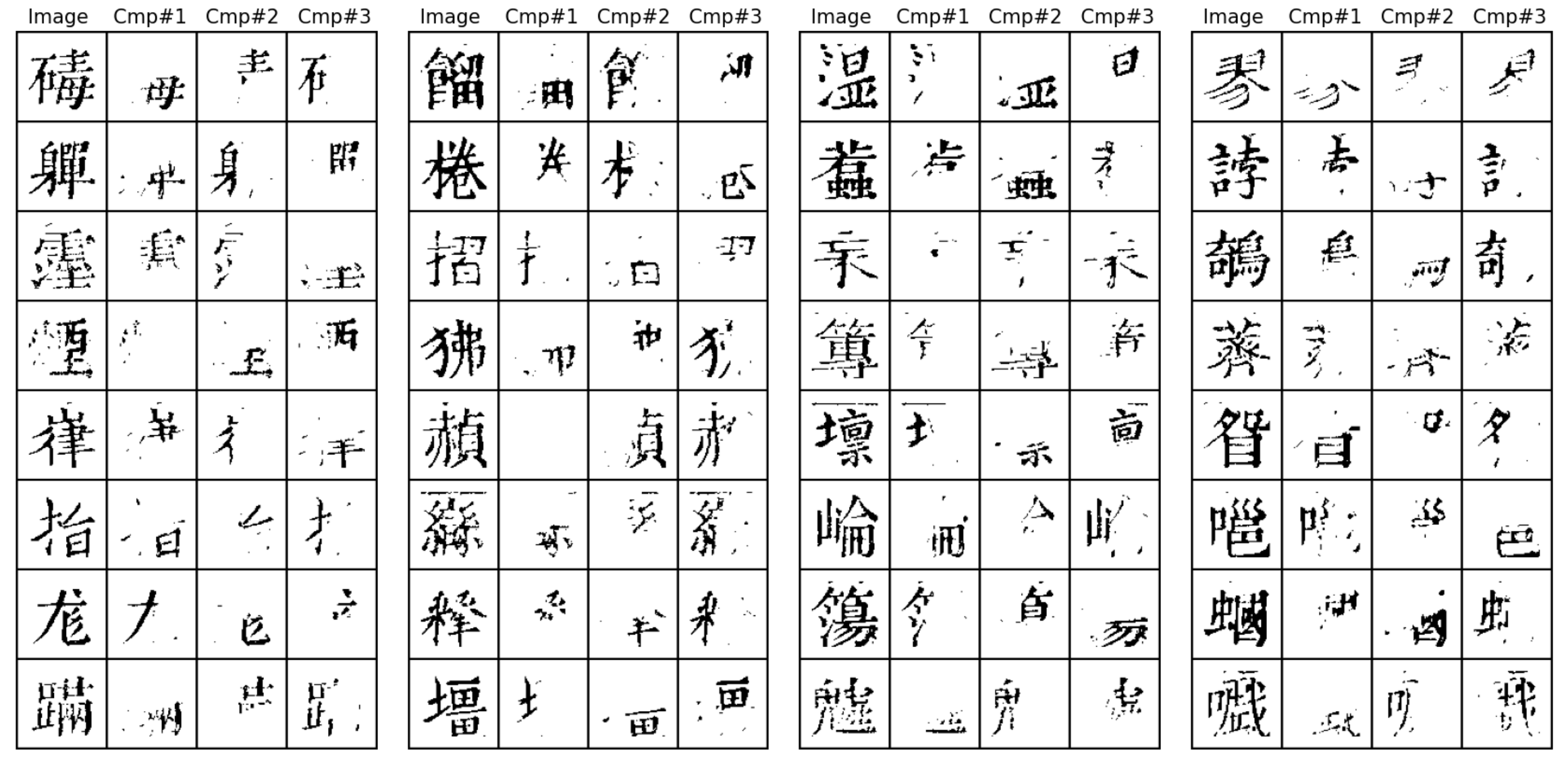}
      \caption{Historical Document}
      \label{fig:appendix-comp-vis-hwdb}
  \end{subfigure}
  \caption{Visualization of the compositional latent components on the different datasets}
  \label{fig:appendix-comp-vis}
\end{figure*}

To further display the compositional latent components learned by CoLa, Figure \ref{fig:appendix-comp-vis} provides additional visualization results on three different datasets: (a) Printed, (b) HWDB, and (c) Historical Document. We select input images from the test set of each dataset and visualize their compositional latent components (Cmp\#1-\#3). CoLa can decompose characters into meaningful structures in most cases, despite various character styles across the datasets. On images from historical documents (Figure \ref{fig:appendix-comp-vis}c), CoLa still successfully extracts compositional latent components, even if some images are incomplete or partly ambiguous. These results demonstrate that CoLa can learn interpretable components across different styles of Chinese characters and handle low-quality images.

\subsection{Additional Results of Zero-shot CCR}

\begin{table}[!t]
    \centering
    \caption{Accuracy (\%) of character zero-shot CCR on HWDB and Printed.}
    \begin{tabular}{l ccccc}
    \toprule
    \multirow{2}{*}{Datasets}  & \multicolumn{5}{c}{HWDB (Character Zero-shot)} \\ 
    \cmidrule(l){2-6}
    & 500 & 1000 & 1500 & 2000 & 2755 \\
    \midrule
    CoLa & 68.59 $\pm$ 0.20 & 76.58 $\pm$ 0.08 & 79.16 $\pm$ 0.18 & 81.16 $\pm$ 0.14 & 82.71 $\pm$ 0.01 \\
    \midrule
    \multirow{2}{*}{Datasets}  & \multicolumn{5}{c}{Printed (Character Zero-shot)} \\ 
    \cmidrule(l){2-6}
    & 500 & 1000 & 1500 & 2000 & 2755 \\
    \midrule
    CoLa & 78.10 $\pm$ 0.04 & 85.38 $\pm$ 0.07 & 90.32 $\pm$ 0.04 & 93.26 $\pm$ 0.03 & 92.70 $\pm$ 0.07 \\
    \bottomrule
    \end{tabular}
    \label{tab:appendix-czs}
 \end{table}

 \begin{table}[!t]
    \centering
    \caption{Accuracy (\%) of radical zero-shot CCR on HWDB and Printed.}
    \begin{tabular}{l ccccc ccccc}
    \toprule
    \multirow{2}{*}{Datasets}  & \multicolumn{5}{c}{HWDB (Radical Zero-shot)} \\ 
    \cmidrule(l){2-6}
    & 50 & 40 & 30 & 20 & 10 \\
    \midrule
    Ours & 70.40 $\pm$ 0.18 & 74.80 $\pm$ 0.42 & 77.01 $\pm$ 0.12 & 80.64 $\pm$ 0.18 & 75.78 $\pm$ 0.28 \\
    \midrule
    \multirow{2}{*}{Datasets}  & \multicolumn{5}{c}{Printed (Radical Zero-shot)} \\ 
    \cmidrule(l){2-6}
    & 50 & 40 & 30 & 20 & 10 \\
    \midrule
    Ours & 82.23 $\pm$ 0.04 & 84.48 $\pm$ 0.25 & 82.20 $\pm$ 0.01 & 92.12 $\pm$ 0.02 & 94.81 $\pm$ 0.04 \\
    \bottomrule
    \end{tabular}
    \label{tab:appendix-rzs}
 \end{table}

 \begin{table}[!t]
    \centering
    \caption{Accuracy (\%) of zero-shot CCR on Historical Document.}
    \begin{tabular}{l c}
    \toprule
    Models & CoLa \\
    \midrule
    Accuracy & 57.37 $\pm$ 0.10 \\
    \bottomrule
    \end{tabular}
    \label{tab:appendix-hd}
 \end{table}

The experiment results of CoLa include uncertainty, since it samples the latent variables and decomposition order from probability distributions. In Tables \ref{tab:appendix-czs}, \ref{tab:appendix-rzs} and \ref{tab:appendix-hd}, we provide the standard deviation and the average accuracy, where three test trials are conducted for each configuration. The results show that the uncertainty of CoLa does not have a significant influence on its performance.

\subsection{Influence of the Component Order}

\begin{figure*}[!h]
  \centering
  \begin{subfigure}[b]{\textwidth}
      \centering
      \includegraphics[width=\textwidth]{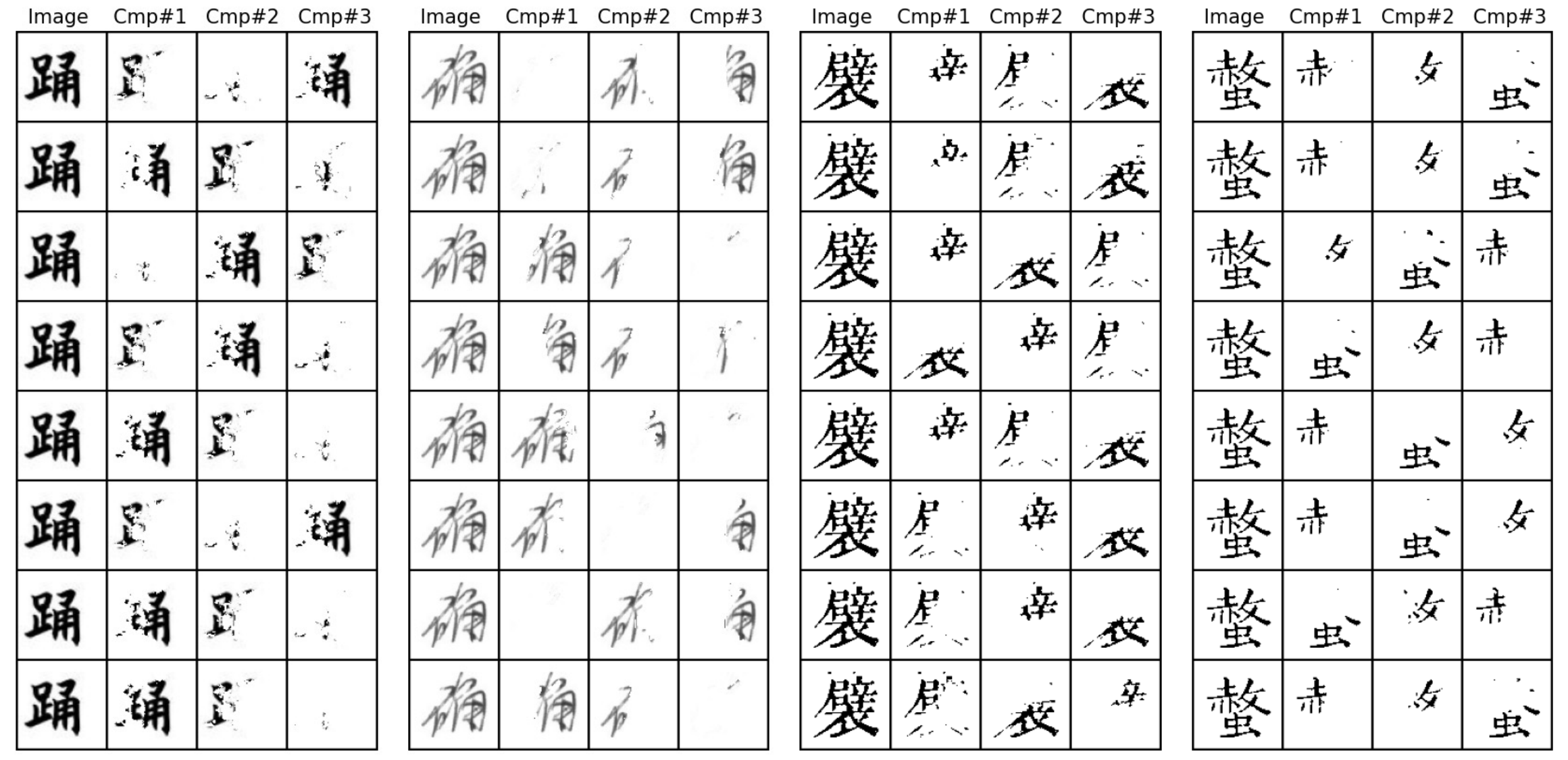}
      \caption{Random initialization}
      \label{fig:appendix-slot-init-random}
  \end{subfigure}
  \vskip 0.2cm
  \vfill
  \begin{subfigure}[b]{\textwidth}
      \centering
      \includegraphics[width=\textwidth]{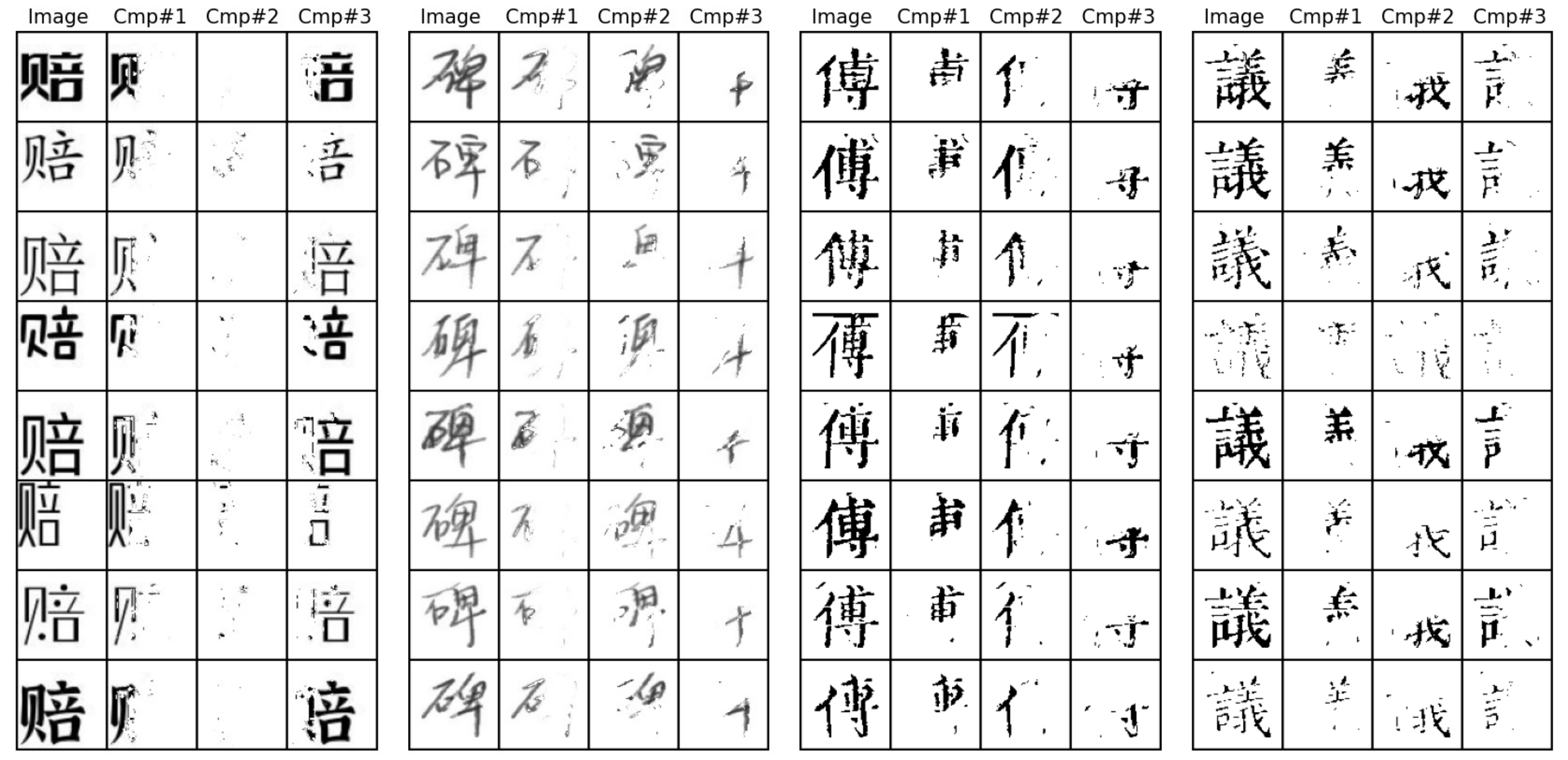}
      \caption{Fix initialization}
      \label{fig:appendix-slot-init-fix}
  \end{subfigure}
  \caption{\textbf{The compositional latent components learned with different orders.} (a) Random initialization. The components in each panel are learned with randomly sampled $\boldsymbol{\epsilon}$. (b) Fix initialization. The components in each panel are learned using a fixed $\boldsymbol{\epsilon}$.}
  \label{fig:appendix-slot-init}
\end{figure*}

This experiment investigates how the decomposition order $\boldsymbol{\epsilon}$ influences the learning of compositional latent representations. Figure \ref{fig:appendix-slot-init} compares the components learned in different ways of initialization: (a) using random initialization, where CoLa samples $\boldsymbol{\epsilon}$ from the Gaussian distribution randomly for each example; (b) using fixed initialization, where a fixed order is used for all examples. For each initialization method, we visualize four groups of character images with their compositional latent components (Cmp\#1-\#3). In Figure \ref{fig:appendix-slot-init}a, the components learned with random initialization have similar decomposition, but the component order varies across examples. With a fixed order (In Figure \ref{fig:appendix-slot-init}b), CoLa assigns semantically or spatially similar components in the same order for one character. These results demonstrate that CoLa’s decomposition process is controlled by the order $\boldsymbol{\epsilon}$, which works by initializing the latent component with $\boldsymbol{\epsilon}$ before input into the slot attention module.

\subsection{Cross-dataset Evaluation}

\begin{figure*}[!h]
  \centering
  \begin{subfigure}[b]{\textwidth}
      \centering
      \includegraphics[width=0.85\textwidth]{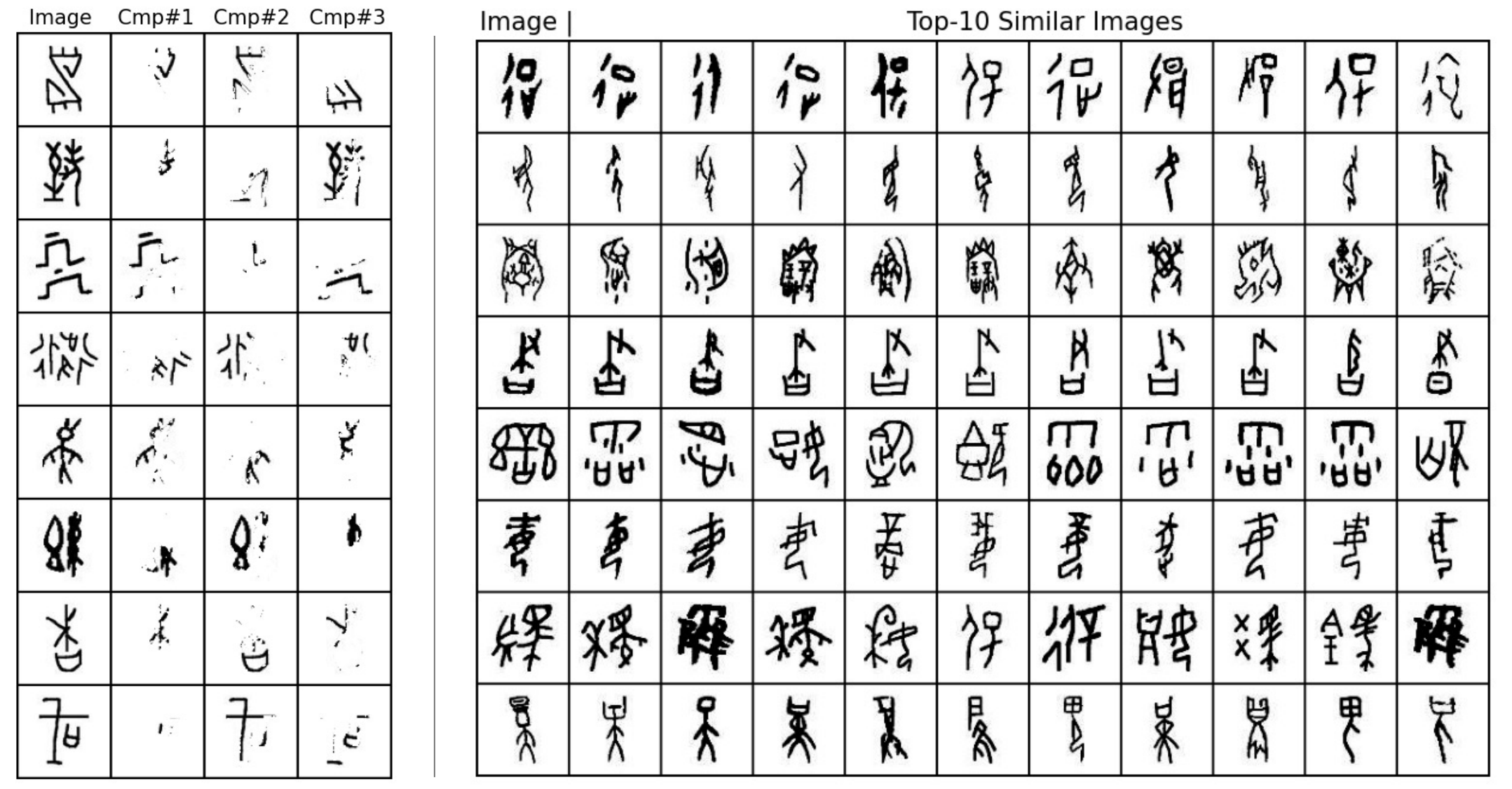}
      \caption{OBC}
      \label{fig:appendix-cross-dataset-obc}
  \end{subfigure}
  \vskip 0.2cm
  \vfill
  \begin{subfigure}[b]{\textwidth}
      \centering
      \includegraphics[width=0.85\textwidth]{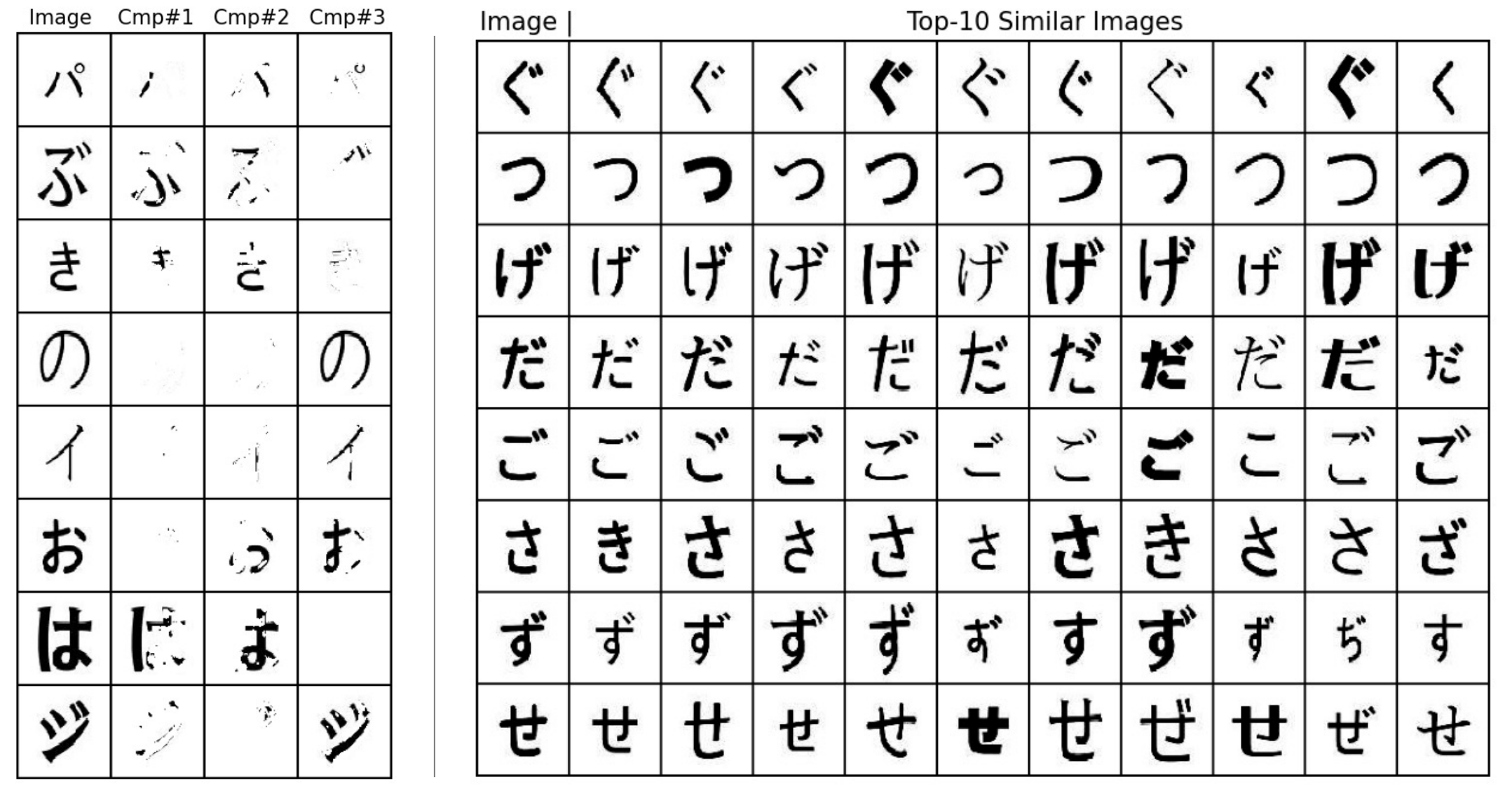}
      \caption{Japanese}
      \label{fig:appendix-cross-dataset-ja}
  \end{subfigure}
  \vskip 0.2cm
  \vfill
  \begin{subfigure}[b]{\textwidth}
      \centering
      \includegraphics[width=0.85\textwidth]{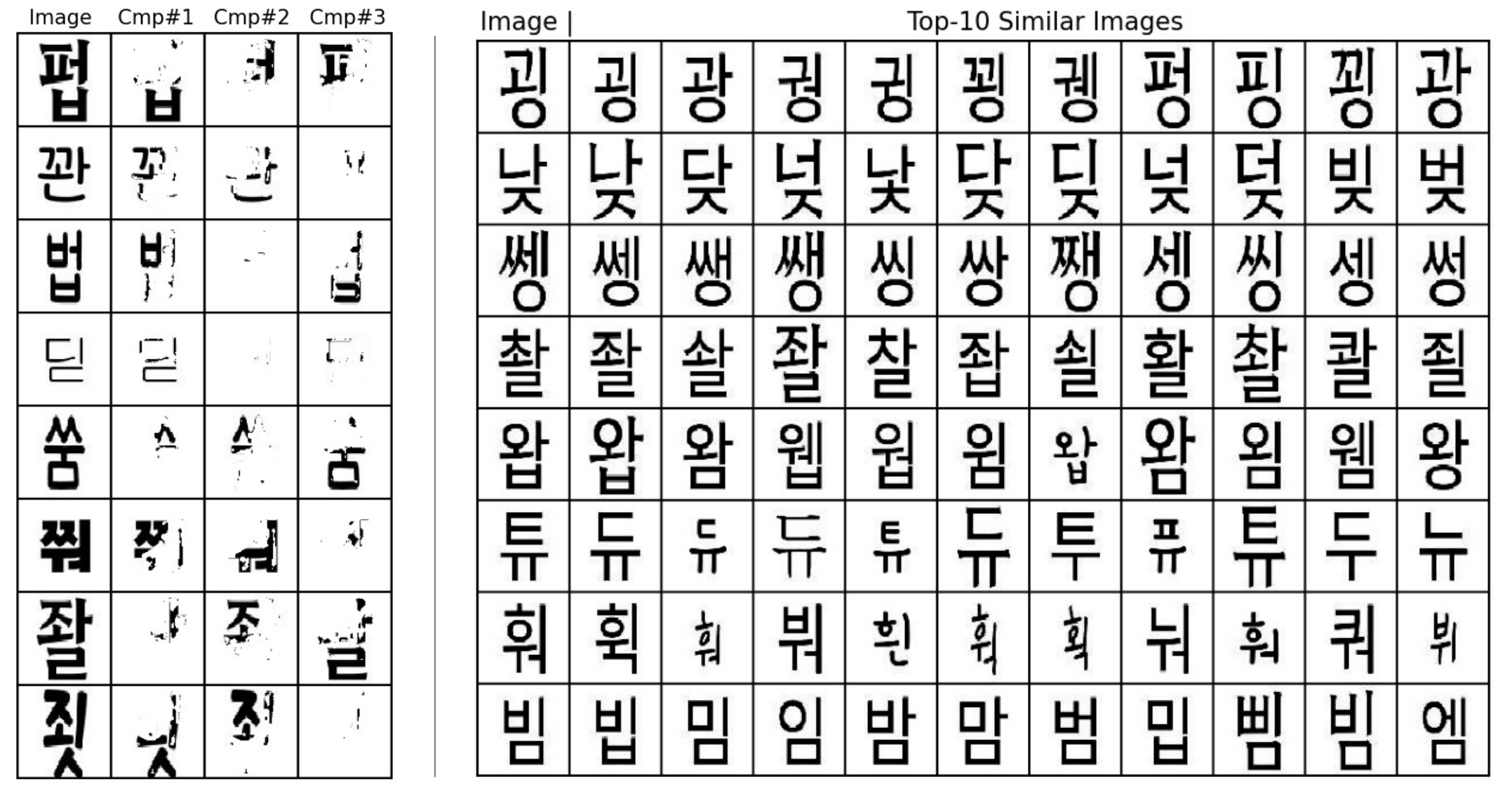}
      \caption{Korean}
      \label{fig:appendix-cross-dataset-kr}
  \end{subfigure}
  \caption{\textbf{The cross-dataset evaluation results.} We evaluate the CoLa trained with historical documents on OBCs, Japanese characters and Korean characters. The left panels are character decomposition results, and the right panels are top-10 similar images of the examples.}
  \label{fig:appendix-cross-dataset}
\end{figure*}

This experiment evaluates the cross-dataset generalization capability of CoLa. Specifically, we aim to assess whether the CoLa trained solely on historical document Chinese characters can transfer its decomposition ability to other types of characters without retraining. To this end, we introduce three datasets with Oracle Bone Characters (OBCs), Japanese characters, and Korean characters. Figure \ref{fig:appendix-cross-dataset} visualizes the decomposition results (Cmp\#1-\#3) and the top-10 similar images retrieved based on the similarity of compositional latent components. For OBCs, despite the large visual gap between modern and oracle bone scripts, CoLa can extract components and retrieve visually and semantically related samples. CoLa can parse distinct components and retrieve similar samples for Japanese and Korean characters. We find that parsing Korean characters is more difficult than Japanese ones, since Korean characters have a larger set of similar components. The experiment confirms that CoLa exhibits the cross-dataset generalization ability and can discover compositional structures in unseen writing systems.

\subsection{Average Inference Time}

To evaluate the computational efficiency of the models, we estimate the per-batch inference time required for predicting the label of input images on a dataset with 3,755 classes of characters. We set the batch size to 32 and compute the average inference time across all batches in the test set during the evaluation phase. As shown in Table \ref{tab:ait}, CoLa demonstrates higher time efficiency compared to previous zero-shot Chinese character recognition (CCR) methods.

\begin{table}[htp]
    \centering
    \small
    \caption{Comparison of average inference time (AIT).}
    \begin{tabular}{l ccccc}
    \toprule
    Methods & Ours & DenseRAN & HDE & SD & CCR-CLIP \\
    \midrule
    AIT(ms) & \textbf{9} & 1666 & 29 & 567 & 14 \\
    \bottomrule
    \end{tabular}
    \label{tab:ait}
 \end{table}

\bibliographyappendix{appendix}
\bibliographystyleappendix{nips}

\newpage

\end{document}